\pdfoutput=1
\documentclass[11pt,a4paper]{article}
\usepackage[hyperref]{acl2021}
\usepackage{times}
\usepackage{latexsym}
\usepackage{amsfonts}
\usepackage{amssymb}
\usepackage{amsmath}
\usepackage{booktabs}
\usepackage{cleveref}
\usepackage{makecell}
\usepackage{multirow}
\usepackage{adjustbox}
\usepackage{scalerel}
\usepackage{xspace}
\usepackage{bbm}
\usepackage{algorithm}
\usepackage{longtable}
\usepackage{arydshln}
\usepackage{tikzlings}
\usepackage[most]{tcolorbox}
\usepackage[noend]{algpseudocode}
\usepackage[font=small,skip=5pt]{caption}

\usepackage{subcaption}
\algnewcommand{\parState}[1]{\State%
    \parbox[t]{\dimexpr\linewidth-\algmargin}{\strut\hangindent=\algorithmicindent \hangafter=1 #1\strut}}

\algrenewcommand\algorithmicindent{1.0em}%

\DeclareMathSymbol{\shortminus}{\mathbin}{AMSa}{"39}
\newcommand\rurl[1]{%
  \href{https://#1}{\nolinkurl{#1}}%
}

\usepackage{cleveref}
\crefname{section}{\S}{\S\S}
\crefname{table}{Tab.}{}
\crefname{figure}{Fig.}{}
\crefname{algorithm}{Alg.}{}
\crefname{equation}{Eq.}{Eq.}
\crefname{appendix}{App.}{}
\crefname{theorem}{Theorem}{}
\crefname{prop}{Proposition}{}
\crefname{cor}{Corollary}{}

\crefname{hypothesis}{Hyp.}{Hypotheses}
\crefformat{section}{\S#2#1#3}

\newcommand{\defeq}{\mathrel{:\mkern-0.25mu=}}

\newcommand{\freq}{\omega}
\newcommand{\stat}{\phi}
\newcommand{\unique}{u}
\newcommand{\calC}{\mathcal{C}}

\newcommand{\defn}[1]{\textbf{#1}}

\newcommand{\vv}{\mathbf{v}}
\newcommand{\yy}{\mathbf{y}}

\newcommand{\calY}{\mathcal{Y}}

\newcommand{\calS}{\mathcal{S}}
\newcommand{\calL}{\mathcal{L}}

\newcommand{\vtheta}{{\boldsymbol \theta}}
\newcommand{\ptheta}{p_{\scaleto{\vtheta}{4pt}}}

\newcommand{\vocab}{\mathcal{V}}
\newcommand{\eos}{\textsc{eos}\xspace}
\newcommand{\bos}{\textsc{bos}\xspace}
\newcommand{\mathcheck}[1]{#1}

\newcommand{\tabitem}{~~\llap{\textbullet}~~}
\definecolor{darkpurple}{RGB}{102,51,153}
\definecolor{BrickRed}{RGB}{155,17,17}
\definecolor{ForestGreen}{RGB}{31,125,12}
\newcommand{\dd}[1]{\textcolor{darkpurple}{\footnotesize $\pm#1$}}

\usepackage[disable]{todonotes}
\makeatletter
\newcommand*\iftodonotes{\if@todonotes@disabled\expandafter\@secondoftwo\else\expandafter\@firstoftwo\fi}  
\makeatother

\newcommand{\note}[4][]{\todo[author=#2,color=#3,size=\scriptsize,fancyline,caption={},#1]{#4}} 
\newcommand{\ryan}[2][]{\note[#1]{ryan}{violet!40}{#2}}
\newcommand{\clara}[2][]{\note[#1]{clara}{orange}{#2}}

\usepackage{microtype}

\aclfinalcopy 
\def\aclpaperid{2031} 

\newcommand{\ucambridge}{\scaleto{\text{\tikz\bear;}}{10pt}}
\newcommand{\ethz}{\scaleto{\text{\tikz\koala;}}{10pt}}

\title{Language Model Evaluation Beyond Perplexity }

\author{Clara Meister$^{\ethz}$ \hspace{3mm} Ryan Cotterell$^{\ethz,\ucambridge}$ \\
   $^{\ethz}$ETH Zürich \hspace{2mm} $^{\ucambridge}$University of Cambridge\\
   \texttt{\{first.last\}@inf.ethz.ch}
  }

\date{}

\begin{document}
\maketitle
\begin{abstract}
We propose an alternate approach to quantifying how well language models learn natural language: we ask how well they match the \emph{statistical tendencies} of natural language. 
To answer this question, we analyze whether text generated from language models exhibits the statistical tendencies present in the human-generated text on which they were trained.
We provide a framework---paired with significance tests---for evaluating the fit of language models to these trends.
We find that neural language models appear to learn only a subset of the tendencies considered, but align much more closely with empirical trends than proposed theoretical distributions (when present).\ryan{What is the empirical--theoretical split?} Further, the fit to different distributions is highly-dependent on both model architecture and generation strategy. As concrete examples, text generated under the nucleus sampling scheme adheres more closely to the type--token relationship of natural language than text produced using standard ancestral sampling; text from LSTMs reflects the natural language distributions over length, stopwords, and symbols surprisingly well.  
\end{abstract}


\section{Introduction}

Neural language models\footnote{In this work, we do \emph{not} use the term language model to refer to cloze language models such as BERT \cite{devlin-etal-2019-bert}, which do not give us a distribution over strings.} have become shockingly good at modeling
natural language data in recent years \cite{merity2017regularizing, xlm, radford2019language}. 
Thus, to test just how well neural language models capture language NLP researchers have started to look beyond standard evaluation metrics such as perplexity, endeavoring to understand which underlying attributes of human language these models are learning.
To this end, a nascent literature has emerged that focuses on probing language models \cite{belinkov-glass-2019-analysis}, i.e., determining whether models encode linguistic phenomena. 
For the most part, these works have been limited to analyses of sentence-level phenomenon, such as subject--verb agreement \cite{gulordava-etal-2018-colorless} and garden path effects \cite{van-schijndel-linzen-2018-neural} among a myriad of other properties \cite[][\textit{inter alia}]{blevins-etal-2018-deep,chowdhury-zamparelli-2018-rnn}.\looseness=-1

\begin{figure}
\centering
    \includegraphics[width=\linewidth]{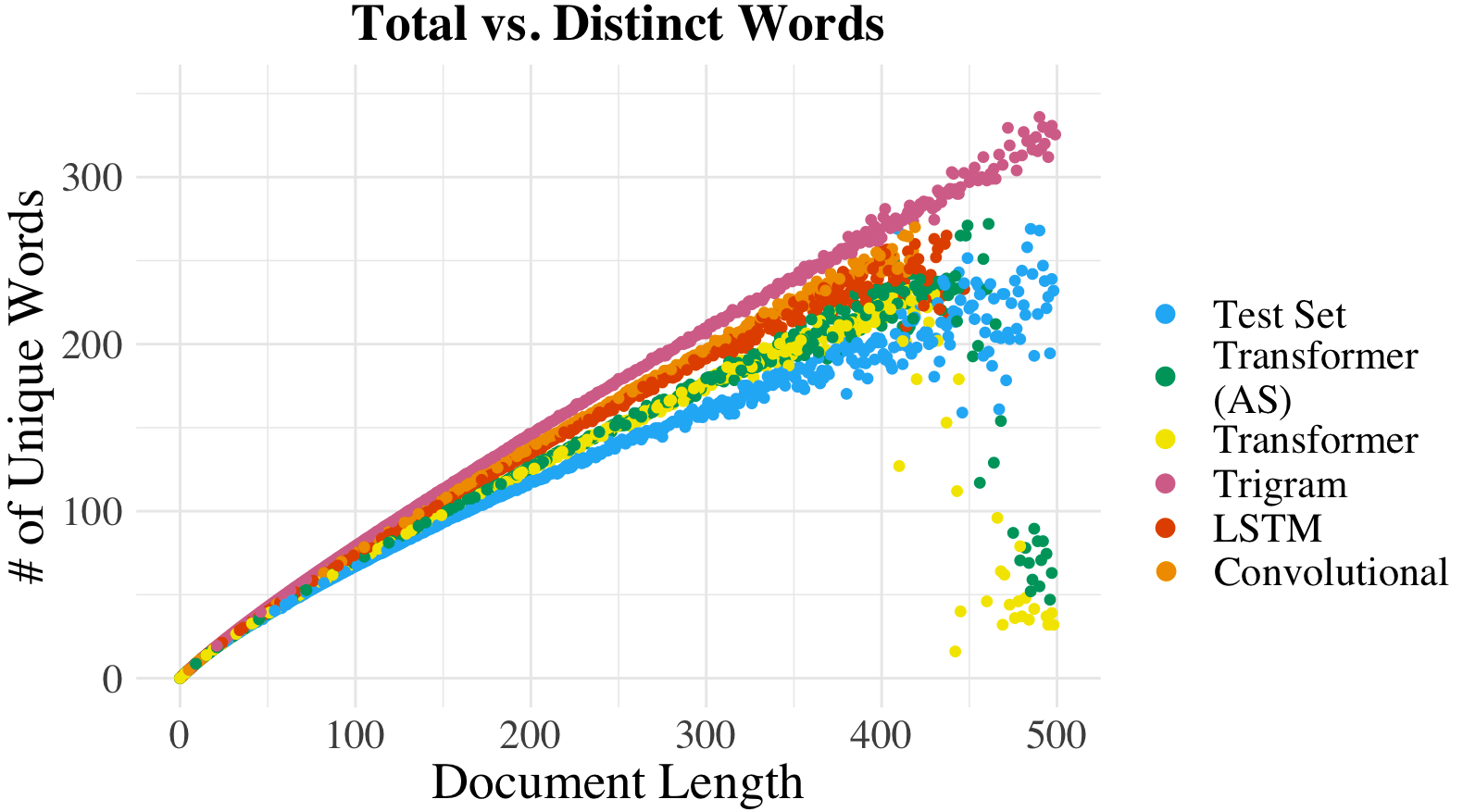}
  \setlength{\belowcaptionskip}{-10pt}
    \caption{Average number of unique words vs. document length, i.e., type--token, in text sampled from language models. Values from models' test set are plotted for reference.\looseness=-1  }\label{fig:heapslaw}
\end{figure}

In this work, we attempt to understand which macro-level phenomena of human language today's language models reflect. That is, we pose the question:
\emph{Do neural language models exhibit the statistical tendencies of human language?}
Phenomena that can be measured at this level provide an alternate view of a model's comprehension; for example,
rather than exploring whether morphological agreement is captured, we look at whether our models learn the trends across a corpus as a whole, e.g., the token rank--frequency (Zipf's) relationship. In comparison to standard probing techniques, this framework does not require we know \textit{a priori} how linguistic phenomena should manifest themselves. That is, when there is no law stating the theoretical tendencies of an attribute of natural language or we have reason to believe our language domain does not follow such a law, we can use the statistical tendencies present in \emph{empirical} data as our baseline. This characteristic both allows us to assess a model's fit to highly corpus-dependent distributions---like the length distribution---and mitigates the biases introduced by our own preconceptions regarding properties of natural language.\footnote{Such biases are naturally introduced by many probing techniques that e.g., draw conclusions from hand-constructed challenge tasks.}\looseness=-1

More concretely, our paper describes an experimental design and accompanying hypothesis tests to determine precisely whether text generated from language models follows the same empirical trends as human language. 
Our experiments reveal that adherence to natural language tendencies varies widely with both model architecture and generation strategy, e.g., \cref{fig:heapslaw} shows varying degrees of adherence to the empirical type--token relationship, an artifact that perplexity alone could not reveal. Our findings suggest this framework is a valuable tool for gaining a deeper understanding of where today's language models are succeeding and failing at capturing human language.\looseness=-1

\section{Language Models}
Language models are probability distributions over natural language sentences.
We define the support of a language model $\ptheta$ with parameters $\vtheta$ as
\begin{equation}
    \calY \defeq \{\bos \circ \vv \circ \eos \mid \vv \in \vocab^*  \}
\end{equation}
\noindent where $\vocab$ is the model's vocabulary and tokens \eos and \bos demarcate the beginning and end of a string, respectively, and $\vocab^*$ is the Kleene closure of $\vocab$. 
In this paper, we term vocabularies consisting of words \textbf{closed} and those consisting of BPE tokens \cite{sennrich-etal-2016-neural} \textbf{open}.

In the case when $\ptheta$ is locally normalized, which is the predominant case for language models, $\ptheta$ is defined as the product of probability distributions:
\begin{equation}\label{eq:joint}
    p_\vtheta(\yy) = \prod_{t=1}^{|\yy|}p_\vtheta(y_t \mid \yy_{<t})
\end{equation}
\noindent where each $\ptheta(\cdot\! \mid \!\yy_{<t})$ is a distribution with support over \mathcheck{$\bar\vocab \defeq \vocab \cup \{\eos\}$ and $\yy_{<1} = y_0  \defeq \bos$}.
To estimate model parameters $\vtheta$, one typically optimizes the log-likelihood function over a corpus $\calC_{\mathrm{train}}$:\looseness=-1
\mathcheck{
\begin{equation}\label{eq:lik}
    \calL (\vtheta \mid \calC_{\mathrm{train}}) = \sum_{\yy \in \calC_{\mathrm{train}}}\log \ptheta (\yy)
\end{equation}}
\noindent where we call each string $\yy$ a \textbf{document}.\clara{add in test/dev}
To determine the goodness of fit of a model to the empirical distribution (defined by $\calC_{\mathrm{train}}$), it is standard practice to measure perplexity on a held-out dataset, which is simply a monotonic function of average (per token) log-likelihood under that model. 
While low perplexity on an evaluation set undoubtedly reflects some level of fit to natural language, it does not give us a fine-grained view of which linguistic attributes a model has learned.

\section{Statistical Tendencies of Language}\label{sec:laws}
Human languages are thought to exhibit statistical tendencies, several of which are explicitly quantified by laws \cite{Altmann_2016}. 
In this section, we review a subset of these distributions--both with and without well-established forms---over which we subsequently perform analyses.\looseness=-1
\subsection{Classical Laws}\label{sec:classic-laws}

\paragraph{Rank--Frequency.}
\citeauthor{zipf1949human}'s law (\citeyear{zipf1949human}), otherwise known as the rank--frequency law, states that the frequency of a word in a corpus decays exponentially in the frequency rank of that word, i.e.,  the frequency $\freq(\cdot)$ of the $k^{\text{th}}$ most frequent word $w_k$ follows the power-law distribution: \mathcheck{$\freq(w_k) \propto k^{-s}$}. When fit to natural language text, the free parameter $s$ is typically close to $1$.
Zipf's law also has a probabilistic interpretation: the marginal probability that a random word in our corpus takes on the value of the$k^{\text{th}}$ most frequent can be expressed as\ryan{Shouldn't $W$ be subscripted as well?}\clara{but W is the word RV. $w_k$ is the particular word of kth rank}
\mathcheck{
\begin{equation}\label{eq:zipf}
   p_{\mathrm{zipf}}(W = w_k) = \frac{1}{\zeta(s)} k^{-s} 
\end{equation}}
\noindent where \mathcheck{$\zeta(s) = 1/\sum_{k=1}^\infty k^{-s}$} is the normalizing constant of our probability mass function (pmf). 
The adherence of language to Zipf's law has been widely studied and is considered one of the canonical laws of quantitative linguistics \cite{baroni2009distributions,e12071743,validation}.\looseness=-1 

Estimating $s$ from an observed set of rank--frequency pairs can be done using standard estimation techniques. Here we use the maximum-likelihood estimate\footnote{Derivation in \cref{app:mle}. We may also estimate $s$ using, e.g., least squares over the original or log--log transform of our distribution. However, it has been empirically observed that least-squares estimates under this paradigm are not reliable \cite{power_law} and further, directly incorporate assumptions that contradict power law behavior \cite{schluter:hal-02880544}.} (MLE), employing numerical optimization to solve for $s$ since the MLE of the discrete power law lacks a closed form solution.\looseness=-1

\paragraph{Type--Token.}\label{sec:heaps}
Heaps' law \cite{heaps}, also known as the type--token relationship, states that the number of additional unique tokens (i.e., number of types) in a document diminishes as its length increases. Formally, we can express the expected number of types $\unique(\cdot)$ as a function of the length $l(\cdot)$ of the string $\yy$ via the relationship \mathcheck{$\unique(\yy) \propto l(\yy)^\beta$} where $\beta < 1$ is a free parameter. Types may be, e.g., unigrams or bigrams.\looseness=-1

The above formulation of Heaps' law lacks an obvious probabilistic interpretation. However, if we frame Heaps' law as modeling the \emph{expected value} of the number of types for any given length document, then we can model the relation as a Poisson process, where the marginal distribution over document length follows Heaps' proposed power law. Specifically, we model the number of types for a document of a given length as a non-homogeneous Poisson process \cite[NHPP;][]{ross1996stochastic} where our rate parameter $\lambda(l(\yy))$ is  Heaps' power law relation. The probability that there are $k$ types in a document of length $t$ is then\looseness=-1
\mathcheck{
\begin{equation}\label{eq:nhpp}
    p_\mathrm{heaps}(u(\yy_{\leq t}) = k) = \frac{\lambda(t)^k}{k!}\exp(-\lambda(t))
\end{equation}}
\noindent for \mathcheck{$\lambda(l(\yy)) = \alpha\cdot l(\yy)^\beta$}. Similarly to \cref{eq:zipf}, we can fit parameters $\alpha,\beta$ using MLE (see \cref{app:mle}).\looseness=-1


\subsection{Other Tendencies}
Natural language has other quantifiable distributions, e.g., over document length or unigrams. 
While there may not exist well-established laws for the behavior of these (often highly corpus-dependent) distributions, we can observe their \emph{empirical} distributions w.r.t. a corpus. We review a few here and leave the exploration of others to future work.\looseness=-1

\paragraph{Length.} Using notation from earlier, we estimate the pmf of the distribution over the length of documents in a corpus $\calC$ as 
\mathcheck{
\begin{equation}\label{eq:length}
    \hat p_{l}(l(\yy) = k) \propto \sum_{\mathbf{\yy} \in \calC}\mathbbm{1}\{l(\yy) = k\}
\end{equation}}\clara{think it matters that the subscripts for p and mu are different?}\ryan{Yes. I would make it consistent.}\clara{I made all of these other distributions $\hat p$}\ryan{That's good.}
\noindent We can additionally compute statistics of this distribution, such as sample mean:\ryan{Is this correct? Seems off to me.} \mathcheck{$\hat\mu_{\scaleto{l}{4pt}}(\calC) = 1/|\calC|\sum_{\yy \in \calC} l(\yy)$}.

\paragraph{Unigram.}
Notably, the rank--frequency law of \cref{sec:classic-laws} leaves the categorical distribution over words unspecified, i.e., it defines the frequency for the$k^{\text{th}}$ ranked word without specifying the word itself. In order to make explicit comparisons, we define the unigram distribution w.r.t. corpus $\calC$ as\looseness=-1
\begin{equation}\label{eq:uni}
    \hat p_{\mathrm{uni}}(w) \propto \sum_{w' \in \calC}\mathbbm{1}\{w' = w\}
\end{equation}

\paragraph{Stopwords and Symbols.}
Certain percentages of words in a string consist of either symbols, i.e., numbers and punctuation, or stopwords, i.e., common words such as ``that'' or ``so'' that primarily serve a syntactic function. We can model this percentage as a (continuous) random variable $S$ and estimate its probability density function (pdf) as
\begin{align}\label{eq:stop}
    &\hat p_{\mathrm{stop}}(s < S \leq s + \delta) \\
    &\quad \quad \propto \sum_{\mathbf{\yy} \in \calC}\mathbbm{1}\Big\{\frac{\#\mathrm{stop}(\yy)}{l(\yy)} \in (s, s+\delta] \Big\} \nonumber
\end{align}
\noindent The pdf for symbols is defined similarly. As with our length distribution, we can compute the means $\hat\mu_\mathrm{stop},\hat\mu_\mathrm{sym}$ of these distributions. 

\section{Statistical Distances}\label{sec:tests}
In this work, we aim to quantify the degree to which the linguistic distributions of text generated from language models match---or differ from---those of natural language. To this end, we propose the use of several probability metrics \cite{prob_metrics1,prob_metrics} as our notion of statistical distance.\footnote{Some of these metrics are formally pseudo-distances, as they are not necessarily symmetric. } For each of these metrics, we present \emph{nonparametric} statistical significance tests, i.e.,\clara{not a precise def of nonparametric tests, should we } tests that may be used when the underlying distribution of observed data is not known.\looseness=-1

\subsection{Primary Metrics}\label{sec:perm}
Perhaps the simplest method for measuring the distance between two random variables is through differences in expectations, e.g., means or variances. (Semi-)distances of this nature are formally called \defn{primary metrics}. To estimate this distance, we can use observations from random samples $\calS_1$ and $\calS_2$, e.g., \mathcheck{$\mu_1 - \mu_2 \approx \stat(\calS_1, \calS_2) = \hat\mu (\calS_1) - \hat\mu (\calS_2)$}. 

Observing a value of $\stat(\calS_1, \calS_2) \neq 0$ on its own is not enough to confirm a difference between $\mu_1$ and $\mu_2$; we need to assess whether the observed distance is \emph{significantly} above or below $0$. Formally, our null and alternative hypotheses are:
\begin{tcolorbox}[ams align,colback=blue!5!white,colframe=blue!75!black,fontupper=\linespread{.66}\selectfont]
    \mathrm{H}_0 \!:\,& \stat(\calS_1, \calS_2) = 0 \\\nonumber
    \mathrm{H}_a \!:\,& \stat(\calS_1, \calS_2) \neq 0 \nonumber
\end{tcolorbox}
\noindent In our setting, we typically do not know the theoretical distributions of the random variables generating $\calS_1$ and $\calS_2$, nor of an arbitrary test statistic $\stat$. Consequently, we use resampling techniques to construct the sampling distribution of $\stat(\calS_1, \calS_2)$.\looseness=-1

\paragraph{Permutation Tests.} In a nutshell, a permutation test provides a simple method for constructing the sampling distribution of a test statistic $\phi$ through empirical observations. 
The method uses the value of $\stat$ over all possible rearrangements of the observed data points to represent the distribution of the test statistic under the null hypothesis. Using this distribution, we can determine the probability of observing a value of the test statistic (or a more extreme value), which if low, may give us reason to reject a specific null hypothesis. 
In this work, we only consider statistics $\stat(\cdot, \cdot)$ over two samples. We provide pseudocode for this case in \cref{app:perm}.\footnote{When the number of possible permutations of the data is computationally prohibitive, we may instead use a MC sampling approach, where we sample from the set of possible permutations \cite{perms}. }\looseness=-1

\subsection{Simple Metrics}
Primary metrics provide only a weak measure of the sameness of random variables as they are completely dependent on a single statistic of a distribution.\clara{check or reword} On the other hand, we know a random variable can be completely described by its distribution function. As such, we turn to \defn{simple metrics} of distance between random variables.\looseness=-1

Given cumulative distribution functions (cdfs) $P_1$ and $P_2$ over one-dimensional random variables, the Kolmogorov--Smirnov (KS) metric 
is
\mathcheck{
\begin{equation}\label{eq:ks}
    D(P_1, P_2) = \sup_y |P_1(y) -  P_2(y)|
\end{equation}}
\noindent where $D \in [0,1]$ and $D(\cdot,\cdot)=0$ indicates the distributions are identical. However, not all random variables can be described in terms of a cdf. For categorical distributions where the support of our random variable is not ordinal, the natural counterpart to the  KS metric is the Chi-square distance. This metric has a number of drawbacks (discussed in \cref{app:chi_squared})---primarily that its value can be hard to interpret and so we instead turn to the total variation distance (\textsc{tvd})---a widely used metric of distance between probability distributions. 

Given two pmfs $p_1$ and $p_2$, we define \textsc{tvd} as\looseness=-1
\mathcheck{
\begin{equation}\label{eq:tvd}
    \textsc{tvd}(p_1,p_2) = \sup_y |p_1(y) - p_2(y)|
\end{equation}}
\noindent  where similarly to the  KS metric,  \textsc{tvd} is bounded above by 1 and a value of $0$ indicates identical distributions. 
In our setting, we consider two use cases for the  KS metric and \textsc{tvd}: as distance metrics between an empirical and theoretical distribution (one-sample) and between two empirical distributions (two-sample).
The corresponding hypotheses that we can test with these metrics are:\looseness=-1
\begin{tcolorbox}[ams align,colback=blue!5!white,colframe=blue!75!black,fontupper=\linespread{.7}\selectfont]
\text{O}&\text{ne-Sample Case:}\\
     & \parbox[t]{\linewidth}{\linespread{1}\selectfont $ \mathrm{H}_0$: {\small Sample $\calS$ is drawn from $p$}\\ $\mathrm{H}_a$: {\small Sample $\calS$ is \emph{not} drawn from $p$}} \nonumber\\
\text{T}&\text{wo-Sample Case:}\\\nonumber
     & \parbox[t]{\linewidth}{\linespread{1}\selectfont $ \mathrm{H}_0$: {\small Samples $\calS_1$ and $\calS_2$ are drawn from same $p$}\\ $\mathrm{H}_a$: {\small Samples $\calS_1$ and $\calS_2$ are not drawn from same $p$}} \nonumber
\end{tcolorbox}
\noindent where in the two-sample case, the exact form of $p$ does not need to be known. These hypotheses require the following tests.


\paragraph{The Kolmogorov--Smirov Test.} The KS test \cite{smirnov1948} is a nonparametric goodness-of-fit test originally designed to assess the fit of a continuous cdf to empirically-observed data; the two-sample version tests whether two samples come from the same distribution. The method has since been extended to discrete distributions and is regarded as one of the most widely applicable nonparametric goodness-of-fit tests for comparing two distributions \cite{discrete-gof,validation}. The test uses the  KS metric $D$ as its test statistic; under our null hypothesis, $D$ converges to 0 almost surely in the limit as our number of samples $n \rightarrow \infty$ by the Glivenko–Cantelli theorem.\footnote{Also known as the “fundamental theorem of statistics.”}
We may reject the null hypothesis if our test statistic is greater than the critical value, which is computed based off of our sample size and a desired significance level.\footnote{Under the null hypothesis, our text statistic $D$ follows a Kolmogorov distribution. In the two sample case, the critical value is dependent on the size of both samples.}
\looseness=-1


\begin{figure*}[!htpb]
\centering
    \includegraphics[width=\textwidth]{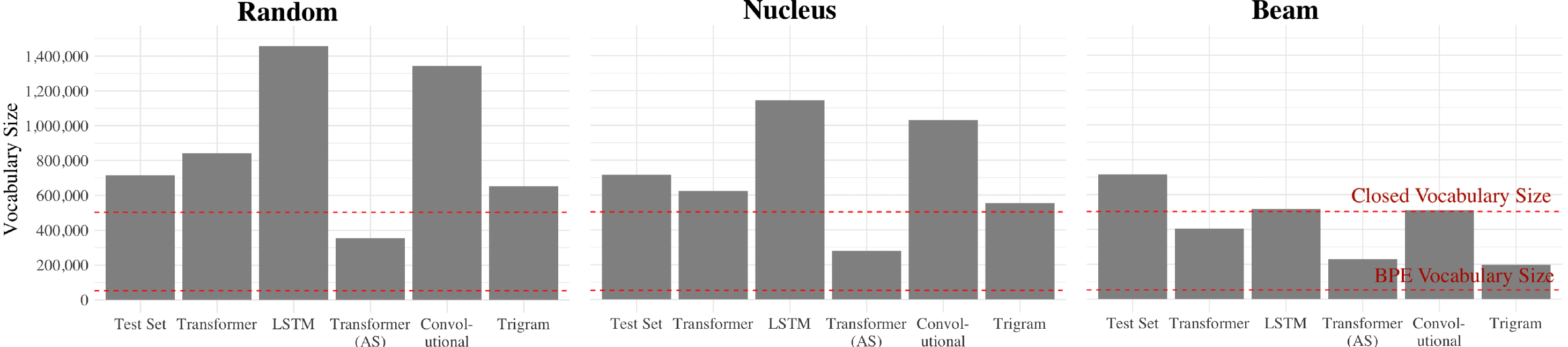}
  \caption{Vocabulary sizes of test set and model-generated samples. Training set (not shown) has vocabulary size of $53.2\mathrm{e}{5}$. Only Transformer (AS) and trigram models have a closed vocabulary; the higher red line is the size of the former.\looseness=-1 }
    \label{fig:vocab}
    \vspace{-5pt}
\end{figure*}
\paragraph{A Test for \textsc{tvd}.}\label{sec:multinomial}
Unlike the  KS metric, we do not have a (theoretical) limiting distribution for \textsc{tvd} between samples from the same distribution that holds for all density functions \cite{tvd_not_c}. However, we can construct this distribution using resampling techniques. Formally, when $\calS_1$ and $\calS_2$ are drawn from the same distribution $p$---where $p$ need not be known---then the test statistic \textsc{tvd}($p_{\scaleto{\calS_1}{4pt}}, p_{\scaleto{\calS_2}{4pt}}$) follows the sampling distribution $\mathcal{Z}_p$, i.e., \mathcheck{$\textsc{tvd}(p_{\scaleto{\calS_1}{4pt}}, p_{\scaleto{\calS_2}{4pt}}) \sim \mathcal{Z}_p$}. The distribution of $\mathcal{Z}_p$ can be computed using permutations of our samples, in the same manner as defined in \cref{sec:perm}. 



\section{Experiments}\label{sec:exp}

We use the above framework to assess the degree to which language models learn various distributions of natural language, i.e., we report metrics outlined in \cref{sec:tests} measured over the distributions and quantities defined in \cref{sec:laws}. 
We compare samples generated from language models to a reserved test set taken from the same corpus as the model's training data. Each set contains 1 million samples.\footnote{Due to our large sample sizes, we should anticipate that our results will almost always be significant, even when effect sizes are trivially small. As such, we will almost assuredly reject our null hypotheses that model-generated samples come from the same distribution as natural language ones. 
While in this light, the presentation of hypothesis tests in \cref{sec:tests} may seem pointless, we provide them for cases where generating many samples for each model setting is computationally prohibitive. } We tokenize all samples using the Moses decoder toolkit \cite{moses}. All text is lower-cased and only complete unigrams are considered, i.e., when BPE is used, only the detokenized unigram is considered. Length of a string is computed as the number of tokens separated by whitespace. Note that when reporting the KS metric ($D$), we always report the metric between (a) an empirical cdf\clara{do we need to clarify how this is done?} computed over the respective model-generated samples and (b) a reference cdf, where $D_{p}$ indicates direct comparison with empirical cdf of the test set. $D_{\ptheta}$ and $D_{\hat p}$ indicate comparison with cdfs of a parametric distribution, whose parameters are estimated on the model and test set, respectively.\looseness=-1 


\paragraph{Natural Language Corpus.}
We use English Wikipedia Dumps,\footnote{\rurl{dumps.wikimedia.org/ }} preprocessing data following the steps used for XLM \cite{xlm} albeit with a $44.7\mathrm{e}{6}$ train--$1\mathrm{e}{4}$ valid--$1\mathrm{e}{6}$ test split. The test set is used in all statistical tests, however, we estimate standard deviations for statistics in \cref{tab:baseline} (in the Appendix) using samples from the training set; see this table for e.g., parameter estimates over test set. 

\begin{figure*}[!htpb]
\centering
    \includegraphics[width=\linewidth]{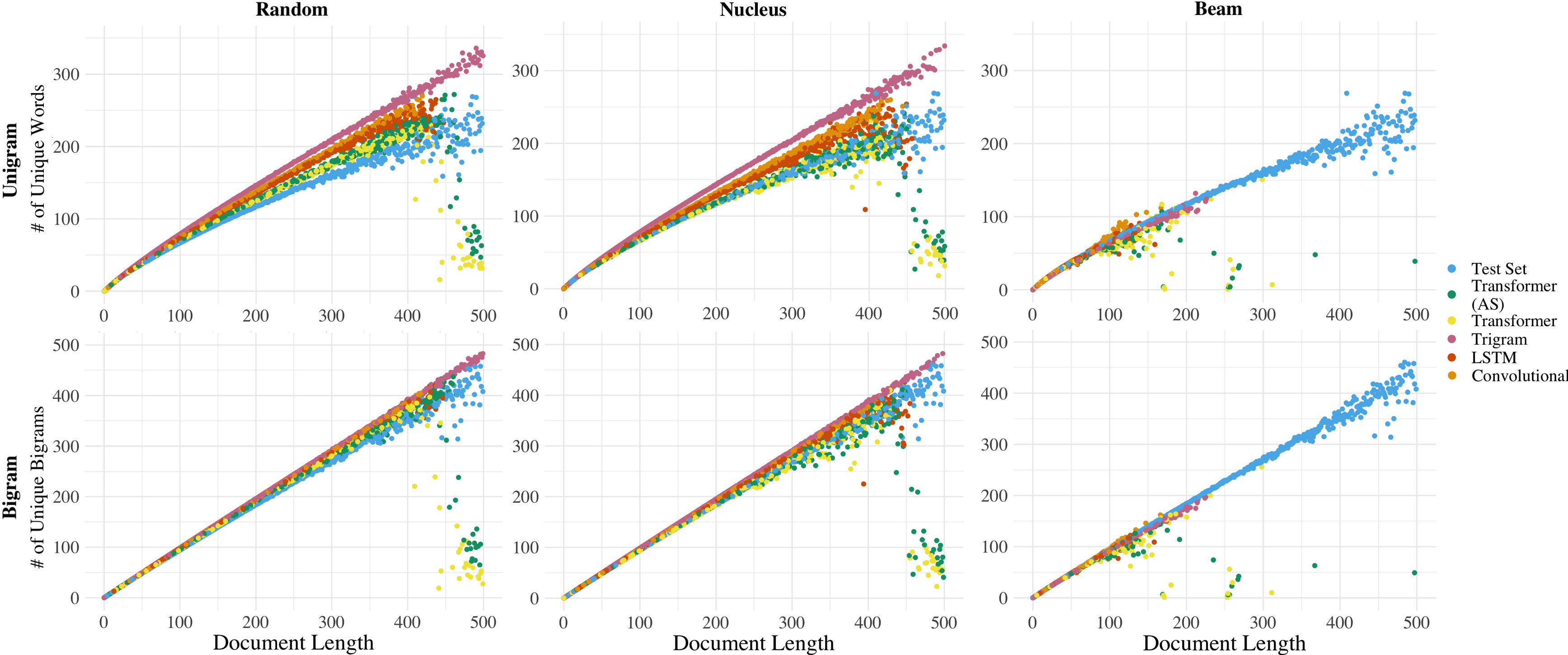}
  \caption{Distinct vs. unique token distributions (unigram and bigram) for test set and text generated from models.\looseness=-1 }
    \label{fig:heapsmain}
    \vspace{-1em}
\end{figure*}

\paragraph{Simulating Corpora from Language Models.} 
Given the distribution $\ptheta$, we may exactly compute statistics and distributions for language models over the entire set $\calY$, weighting examples by the probability assigned to each string; however, doing so is infeasible due to the size of the output space and non-Markovian structure of most neural models.
Rather, we turn to sampling to create a representative set $\calS = \langle\yy^{(1)}, \dots, \yy^{(N)} \rangle$ from $\ptheta$.
We explore three sampling schemes: ancestral random sampling (\textbf{Random}), nucleus sampling (\textbf{Nucleus}), and beam sampling (\textbf{Beam}).\footnote{The latter two sampling designs do not result in samples drawn according to our original $ \ptheta$. As such, the schemes lead to two ``new'' distributions, $\ptheta^{(n)}$ and $\ptheta^{(b)}$, respectively.} 

In ancestral random sampling, $\yy^{(i)}$ are constructed iteratively according to the distribution
\mathcheck{\begin{equation}\label{eq:ancestral}
    y_t^{(i)} \sim \ptheta(\cdot \!\mid \! \yy^{(i)}_{<t})
\end{equation}}
\noindent where $y_0 = \bos$. Under the local normalization scheme of \cref{eq:joint}, sampling according to \cref{eq:ancestral} is equivalent to sampling $\yy^{(i)}$ directly from $\ptheta$. In nucleus sampling, our distribution is truncated to the most probable items covering portion $n \in (0,1]$ of the probability mass. Formally, we now sample \mathcheck{
\begin{equation}\label{eq:nucleus}
    y_t^{(i)} \sim \begin{cases}
\ptheta(\cdot \!\mid \! \yy^{(i)}_{<t})/Z
&\textbf{if }y_t^{(i)}\!\! \in \vocab_n(\ptheta(\cdot \!\mid \!\yy^{(i)}_{<t})) \\
0 & \textbf{otherwise}\end{cases}
\end{equation}}
\noindent where $\vocab_n(p) \subseteq \bar\vocab$ is the smallest subset such that \mathcheck{$\sum_{y \in \vocab_n(p)} p(y) \geq n$ and $Z\defeq\sum_{y\in \vocab_n(p)}p(y)$}. Beam sampling uses \cref{eq:ancestral} as the sampling distribution, but extends a ``beam'' of $k$ sequences at each sampling iteration. I.e., $k$ extensions are sampled from $\ptheta(\cdot \!\mid\! \yy^{(i)}_{<t})$ and the $k$ most probable of the $k^2$ sampled items remain on the beam; note that unlike standard beam search, this is a \emph{stochastic} procedure.\footnote{Note that this is the default sampling scheme for language generation in the \texttt{fairseq} library.} We use a beam size of 5 in all experiments.\looseness=-1





\begin{table}
  \centering
   \adjustbox{max width=\linewidth}{
  \begin{tabular}{lll}
  \toprule
  & \makecell{\textbf{\# params}\\ (millions)} &\makecell{\textbf{Test Set}\\ \textbf{Perplexity}} \\
    \makecell[l]{\textbf{Transformer}} &205 & 23.52\\
    \makecell[l]{\textbf{Transformer} \footnotesize{(adaptive softmax)}} & 315 & 32.66\\
  \makecell[l]{\textbf{Gated Convolutional Network}}& 133 & 48.96\\
  {\textbf{LSTM}  \footnotesize{(3 decoder layers)}} & 59 & 49.29\\
    \bottomrule
  \end{tabular} }
  \caption{Neural language model statistics. }
  \label{tab:models}
  \setlength{\belowcaptionskip}{-20pt}
  \vspace{-1em}
\end{table}
\paragraph{Models.}
We perform our tests on neural models with three different architectures: a transformer \cite{vaswani_attention,adaptive} (only decoder portion), LSTM \cite{hochreiter1997long}, and Convolutional Neural Network \cite{dauphin2017language}. All models are implemented and trained using \texttt{fairseq}.\footnote{\rurl{github.com/pytorch/fairseq/}} We train models on corpora processed both with and without BPE.  We include details for each model in \cref{tab:models}. We additionally estimate a trigram model on the training data; formally, we build a model where the probability of observing token $x\in\bar\vocab$ at position $i$ of the text is estimated\ryan{Are the indices here in the right order?}\clara{yes} as\looseness=-1
\mathcheck{\begin{align}
   &p(x \mid x_{i-2},x_{i-1}) 
    \\
    &\quad=\frac{c(\langle x_{i-2},x_{i-1}, x\rangle)}{\sum_{x'\in\bar\vocab} c(\langle x_{i-2},x_{i-1}, x'\rangle)} \nonumber
\end{align}}
\noindent where $c(\cdot)$ denotes the function counting occurrences of a sequence in some implicit $\mathcal C$. Note that we do not employ smoothing techniques in this model, thus, perplexity over a held-out dataset may diverge and so is not reported in \cref{tab:models}. Vocabulary statistics for each sample are shown in \cref{fig:vocab}. We provide samples of model-generated text in \cref{app:gen_text}.\looseness=-1


\begin{table*}
  \centering
   \adjustbox{max width=0.9\textwidth}{
   \small
  \begin{tabular}{lccc:ccc:ccc|ccc}
  \toprule
  &  \multicolumn{9}{c}{\bf Rank--Frequency} &  \multicolumn{3}{c}{ \bf Unigram} \\
  &  \multicolumn{3}{c}{$D_{\ptheta}$} &  \multicolumn{3}{c}{ $D_{\hat p}$}  &  \multicolumn{3}{c}{$D_{p}$} &  \multicolumn{3}{c}{\textsc{tvd}} \\
 {\bf Model} & ${\bf R}$ &${\bf N}$&${\bf B}$&${\bf R}$&${\bf N}$&${\bf B}$ &${\bf R}$&${\bf N}$ &${\bf B}$ &${\bf R}$&${\bf N}$ &${\bf B}$\\
  \midrule
    Transformer & 0.150 & 0.145& 0.170&0.150 &{\bf\color{ForestGreen}0.142} & 0.170 & {\bf\color{ForestGreen} 3.7e-3} & 0.029 & 0.024& {6.9e-3} & {6.9e-3} & {6.9e-3} \\
    
    Transformer {\footnotesize(AS)} & 0.145& 0.142& 0.150& 0.143  &{\bf\color{ForestGreen}0.142} &{\bf\color{ForestGreen}0.142}& 0.013&0.041 &0.046&0.014 & 0.014 & 0.038\\
    
    CNN & 0.145& 0.142 &0.167& 0.144 & {\bf\color{ForestGreen}0.142}& 0.167& 0.013 & 0.039 & 0.022& 6.9e-3 & {6.9e-3}& {8.6e-3} \\
    
  LSTM & 0.147& 0.143& {\bf\color{BrickRed}0.175}& 0.144 &{\bf\color{ForestGreen}0.142} &{\bf\color{BrickRed}0.178} &0.016 & 0.043 & 0.034  & 3.4e-3& 0.010 &{9.2e-3} \\
  
  Trigram & 0.151 & 0.148& {\bf\color{ForestGreen}0.119}& 0.154 & 0.146 & 0.152 & $4.9\mathrm{e}{\text{-}3}$& {0.020} &{\bf\color{BrickRed}0.251}  &{\bf\color{ForestGreen}2.9e-3} & {3.0e-3}& {\bf\color{BrickRed}0.075}\\
  
  \bottomrule
  \end{tabular} }
  \caption{ KS metrics (lower implies closer fit) between models' empirical cdf and reference cdfs for the rank--frequency relationship. $D_{\ptheta}$ and $D_{\hat p}$ indicate statistical distance from a Zipfian distribution, where parameter $s$ is estimated using the model and test sets, respectively. $D_{p}$ indicates direct comparison with empirical cdf of test set. $p$-values (estimated using Monte Carlo simulations \cite{monte_carlo_KS}) for all  KS metrics are $\ll 0.001$. For the unigram distribution, we report \textsc{tvd} between empirical cdfs of model and test set. All $p$-values are $<0.001$ (see \cref{app:other_results}).\looseness=-1 }
  \label{tab:zipf}
  \setlength{\belowcaptionskip}{-30pt}
\end{table*}

\begin{figure*}
\centering
    \includegraphics[width=\linewidth]{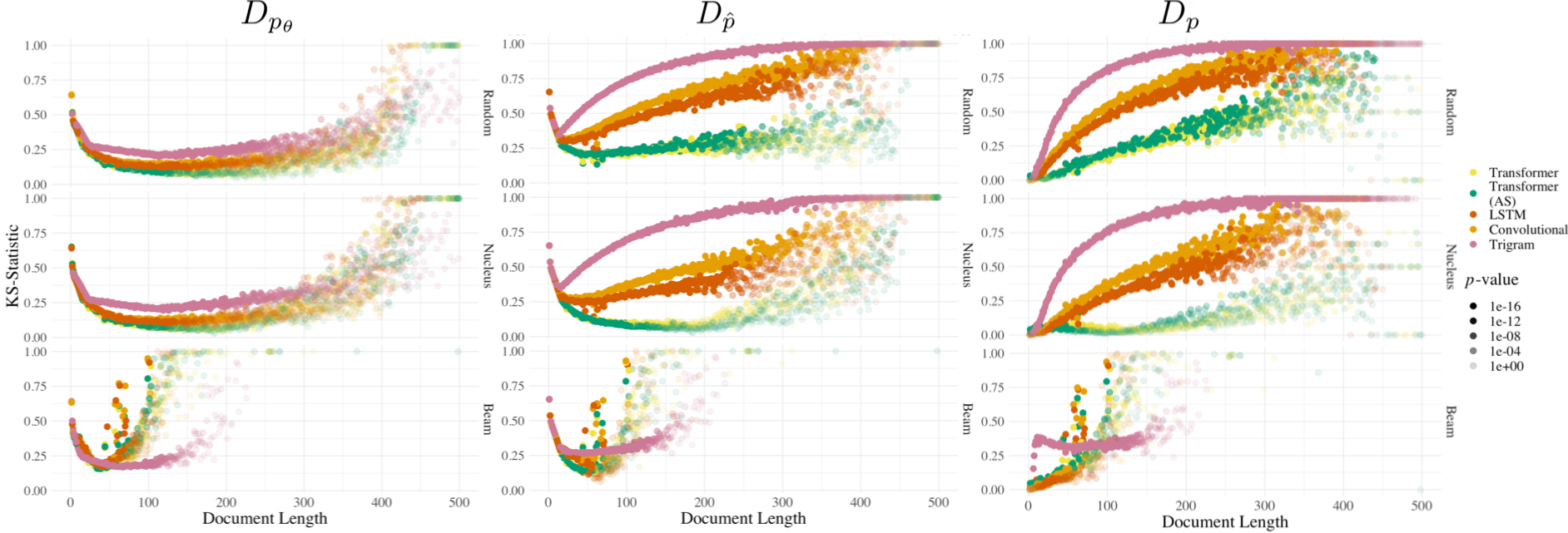}
  
    \caption{ KS metrics (lower implies closer fit) with reference distributions for the type--token relationship as a function of document length. $D_{\ptheta}$ and $D_{\hat p}$ statistical distance from NHPP distribution for params fit to model text and test sets, respectively; $D_{p}$ is computed directly against the empirical cdf of test set. Shading indicates significance of the statistic.\looseness=-1}\label{fig:ks}
    \vspace{-1em}
\end{figure*}
\subsection{Rank--Frequency}\label{exp:zipf}
To understand the rank--frequency relationship implicitly learned by language models---and how it relates to the rank--frequency distribution present in natural language---we compute the three KS metrics previously described: $D_{\ptheta}$, $D_{\hat p}$, and $D_{p}$. Specifically, for the first two values,  we use the cdf of a Zipfian distribution parameterized by $s$ as our reference---where $s$ is estimated using model generated samples or the test set, respectively.\footnote{$s$ is known to vary with the corpus size $|\calC|$ \cite{powers-1998-applications}, however $|\calC|$ is the same for all sets, so this should not affect our analysis.} These metrics give us a sense of how well the rank--frequency distribution under our language models match a Zipfian distribution. Since the power-law behavior of the token rank--frequency distribution is known to fall off at higher ranks \cite{Piantadosi2014ZipfsWF, validation}, we consider solely the first 10,000 ranks in each sample, including when computing $D_{p}$. 
We report these values in  \cref{tab:zipf}. Values of estimates of $s$ and plots of rank--frequency are shown in \cref{app:other_results}.\looseness=-1


Our results indicate that our models' empirical rank--frequency distributions do not adhere very closely to a standard Zipfian distribution (as shown by $D_{\ptheta}$ and $D_{\hat p} \gg 0$),\ryan{what is meant by $\gg$?}\clara{Im trying to say far from zero. Do you have a suggestion?} despite appearing to at a superficial level (see \cref{app:other_results}). However, the same is true for our test ($D_{\hat p}=0.148$), which suggests that our models fit a Zipfian distribution perhaps \emph{no more poorly} than natural language does. Rather, the model produces qualitatively worst text (see \cref{app:gen_text})---a trigram model under the beam sampling generation strategy---follows a power law trend the most closely of any of our samples. On the other hand, the small values of $D_p$ suggest our models learn the \emph{empirical} rank--frequency trends of human text quite well, something that would not be evident by simply looking at adherence to a Zipfian distribution. The combination of these results suggest the limitation of using adherence to Zipf’s law as a gauge for a model’s consistency with natural language. 




\subsection{Type--Token}\label{sec:2}

\cref{fig:heapsmain} shows the type--token trend for all corpora and generation schemes. While most models appear not to follow the same trend as the natural language distribution (as depicted by our test set), we observe that transformers under the nucleus sampling generation scheme match it most closely. Indeed, both models based on the transformer architecture exhibit remarkably similar trends in these experiments, despite having different vocabulary sizes and hyperparameters: both in their generally close fit to the natural language type--token distribution and in their visible fall-off for longer length sequences. The latter observation reveals a deficiency that is seemingly specific to the transformer architecture---one that may be linked to observations in natural language generation tasks. More specifically, we take this as quantitative evidence for recent qualitative observations that  when left to generate lots of text, neural language models based on the transformer architecture tend to babble repetitively \cite{holtzman2019curious, pmlr-v97-cohen19a, eikema2020map}.\looseness=-1


To provide a more mathematically rigorous analysis, we compute KS metrics,\footnote{ \cref{sec:heaps} provides motivation for comparing distributions at individual time steps rather than collectively over time; analyzing \cref{eq:nhpp} for all document lengths simultaneously would not give us a sense of how the power-law fit changes as a function of document length.} again presenting three values: $D_{\ptheta}$, $D_{\hat p}$, and $D_{p}$. In \cref{fig:ks}, we can see that model-generated text follows a NHPP parameterized by Heaps' law moderately well ($D_{\ptheta}$); there are larger divergences at the tails of document length. However, most do not follow an NHPP with the same parameters as our test set ($D_{\hat p}$). Further, in contrast to rank--frequency, the type--token distribution is \emph{more} disparate from the empirical natural language distribution than our parameterized ones, as shown by high values of $D_p$. While both transformers exhibit the closest fit for all document lengths, which is in-line with our observations in \cref{fig:heapsmain}, statistical distance from the natural language distribution for all models and in all settings increases with document length.\looseness=-1




\subsection{Unigram Distribution}
Because we do not have a well-established law dictating the form of the natural language unigram distribution, we compare only empirical pmfs from model-generated samples and the test set directly. 
Further, as the distribution over unigrams is categorical, we employ \textsc{tvd} following \cref{sec:multinomial}. Our results in \cref{tab:zipf} indicate that language models generally capture the unigram distribution quite well. The transformer (AS), which has a closed vocabulary, consistently performs poorly in comparison to other models. While we might speculate this outcome is a result of disparate tails between empirical cdfs---i.e., the part of the distribution over infrequent words, which may have been omitted from the closed vocabulary but could still be generated using BPE---the \textsc{tvd} metric in this setting should generally be robust to tail probabilities.\footnote{We observe this empirically; calculating \textsc{tvd} between distributions truncated to the (union of the) first 1000 ranked unigrams lead to almost the exact same result.} This suggests that BPE (or similar) vocabulary schemes may lead to models that can better fit this natural language distribution. 

\subsection{Length, Stopwords and Symbols}

\begin{table*}
  \centering
  \small
   \adjustbox{max width=0.9\textwidth}{
  \begin{tabular}{llll|lll|lll}
  \toprule
 \multirow{2}{*}{{\bf Model}} &  \multicolumn{3}{c}{\bf Length} &  \multicolumn{3}{c}{ \bf Stopword} &  \multicolumn{3}{c}{ \bf Symbol}\\
  &  Random &  Nucleus & Beam &  Random &  Nucleus & Beam & Random &  Nucleus & Beam\\
  \midrule
    Transformer &0.031& 0.034 & 0.481 &  0.023 &  0.062 &   0.323 & 0.081  & 0.065&  0.205  \\
    Transformer {\footnotesize(AS)} &0.037 & 0.041 & 0.477 & 0.047 & 0.015 & 0.378 & 0.083 &  0.072 & 0.252  \\
    CNN  & 0.034 &  0.051 & 0.491 & 0.036 &0.102 & 0.324 & 0.069 & 0.054 & 0.213\\
  LSTM   &{\bf\color{ForestGreen}0.014} & 0.036& {\bf\color{BrickRed}0.516}& {\bf\color{ForestGreen}0.008}& 0.069.&  0.382 & {\bf\color{ForestGreen}0.037} & 0.048 &  {\bf\color{BrickRed}0.271} \\
  Trigram& 0.093& 0.084 & 0.214 & 0.126 & 0.145 & {\bf\color{BrickRed}0.490} & 0.044 & {\bf\color{ForestGreen}0.037} &  0.061\\

    \bottomrule
  \end{tabular} }
  \caption{ KS metrics ($D_p$) between empirical length, stopword, and symbol distributions of test set and model generated text. $p$-values (estimated using Monte Carlo simulations \cite{monte_carlo_KS}) for all  KS metrics are $\ll 0.001$.}
  \label{tab:stop}
  \vspace{-1em}
\end{table*}

Similarly to the unigram distribution, for length, stopwords and symbols, we compare solely empirical cdfs. 
We use the set of English stopwords defined by NLTK \cite{nltk}. We define the set of symbols as tokens consisting solely of punctuation and numerical values. 
Our results in \cref{tab:stop} demonstrate that our language models---at least when using random and nucleus sampling---mimic these natural language distributions quite well. Notably, text generated from an LSTM using random sampling follows all three distributions the closest of any model, suggesting LSTMs may have an inductive bias that is helpful for capturing these distributions. On the other hand, using beam sampling leads to strong divergence from natural language distributions across the board.  Results for differences in distribution means in the permutation testing framework can be found in \cref{app:other_results}.\looseness=-1

\begin{figure}
\centering
    \includegraphics[width=0.9\linewidth]{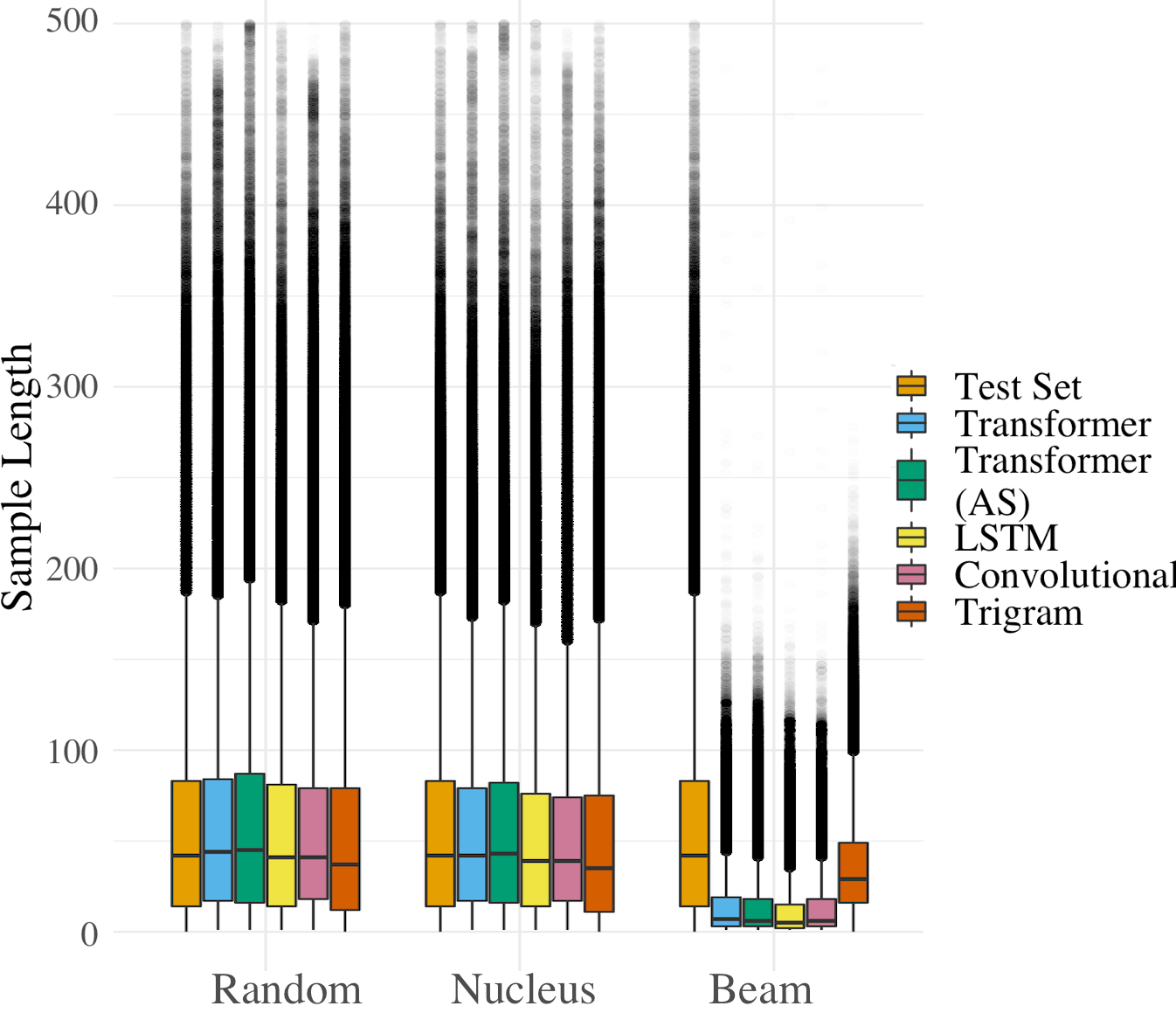}
  \caption{Boxplots showing the distribution of sample length per model and generation scheme. Distribution of test set is repeated in each group for reference.}
  \vspace{-1em}
    \label{fig:lengthbox}
\end{figure}

With respect to the length distribution, these results are perhaps surprising: the local-normalization scheme used by the majority of language generation models (and by those in these experiments) has been claimed to result in models that favor shorter than typical sequences \cite{sountsov-sarawagi-2016-length,murray-chiang-2018-correcting}. The results in \cref{tab:stop,fig:lengthbox} suggest otherwise. Specifically, we see that our models fit the natural language length distribution of our corpus quite closely, in terms of both overall distributions and means (see \cref{app:other_results}). Rather, it appears that the generation strategy may be the cause of prior observations. 
This finding  raises further questions: since models capture the length distribution well, is a language model more likely to produce degenerate text (e.g., repetitions) than the EOS token if only long documents are used in training? We posit that corpus preprocessing should perhaps be more carefully considered in light of these results. 

\subsection{Consistent Trends}
Across results, we observe that text generated using the nucleus sampling decoding scheme often aligns with natural language more closely than text produced using other generation strategies. This suggests that nucleus sampling performs a helpful alteration to a standard distribution learned via MLE, which may in turn provide motivation for recent efforts to employ truncated or sparse probability distributions directly at training time, e.g., truncated loss \cite{kang-hashimoto-2020-improved} or $\alpha$-entmax loss \cite{peters-etal-2019-sparse}.

We additionally observe large discrepancies in both \cref{exp:zipf} and \cref{sec:2} between the results when using empirical natural language cdfs vs. parametric ones. We take this 
as a warning that assumptions about the forms of linguistic distributions---such as the ones employed by challenge tasks in probing---can have significant effects on results.

\section{Related Work}\clara{incorporate \cite{kuncoro-etal-2019-scalable}  and \cite{mueller-etal-2020-cross}? or GLTR and HUSE?}
In the last few years, a number of works have extended language model analysis beyond simple evaluation metrics---like perplexity---in order to understand what attributes of human language these models are learning. Some use task-based approaches, i.e., they design a set of tasks that require a specific subset of linguistic knowledge then evaluate model performance on these tasks \cite[][\textit{inter alia}]{linzen-etal-2016-assessing,gulordava-etal-2018-colorless,tacl_howdoweknow}. 
Others use model-based approaches, where a separate model is trained to perform some auxiliary task on representations learned by the model under test \cite[][\textit{inter alia}]{blevins-etal-2018-deep, giulianelli-etal-2018-hood,sorodoc-etal-2020-probing}. We direct readers to \citet{belinkov-glass-2019-analysis} for a full survey of probing methods.\looseness=-1

These approaches have drawbacks; for example, introducing a secondary model to determine what the original model has learned presents confounding factors  \cite{hewitt-liang-2019-designing}. The designing of auxiliary tasks for assessing linguistic knowledge requires large manual effort and lends itself to implicit bias about how linguistic phenomena should manifest. In contrast, our work allows us to take a hands-off\clara{better word?} approach to analyzing language models. We see the benefit of this in \cref{sec:exp}, where our results \emph{without} an assumed model of statistical tendencies give us a much different sense of which empirical properties of human-generated text our models have learned.\looseness=-1


Our work is closest to that of \citet{takahashi_statistical,takahashi_evaluating} who use model generated text to visually analyze whether language models reflect well-established statistical tendencies. 
In contrast, our work provides a quantitative framework, along with appropriate significance tests,\footnote{In this respect, our work is similar to \citet{dror-etal-2018-hitchhikers}, whom also present statistical tests for use in NLP.} for evaluating distribution fits. 
We additionally assess the fit of language models to our test set directly, rather than solely to established laws. 
Further, our analysis includes different generation strategies, multiple neural architectures, and a wider variety of empirical language distributions.\looseness=-1



\section{Conclusion and Future Directions}
In this work, we present a framework for determining the linguistic properties learned by language models through analysis of statistical trends in generated text. We find that neural language models accurately capture only a subset of natural language distributions and that this subset is highly dependent on both model architecture and generation strategy; no one configuration stands out as capturing all linguistic distributions. 
Ultimately, we see this analysis framework as a means for a more fine-grained evaluation of language models than perplexity alone can provide. Uncovering which linguistic properties language models have learned---and which they have not---should help us to understand both the inductive biases of various models and via which avenues they can still be improved.\looseness=-1

There are a number of important axes of variation that this work does not explore: perhaps most importantly, our results are limited to a single corpora in the English language. A cross-linguistic analysis may reveal whether different model architectures exhibit inductive biases compatible with different languages; observing how these metrics change as a function of corpus size would have implications about the effects of data availability. An exploration of the correlation of these metrics with other quantifications of model performance, such as perplexity or a model’s ability to capture sentence level phenomenon, may help us understand how comprehensive other evaluation metrics are. We leave these analyses as future work.

\section*{Acknowledgements}
We thank Adhi Kuncoro for helpful discussion and feedback in the middle stages of our work and Tiago Pimentel, Jason Wei, and our anonymous reviewers for insightful feedback on the manuscript. We additionally thank B. Bou for his concern.


\bibliography{anthology,acl2020}

\begin{thebibliography}{49}
\expandafter\ifx\csname natexlab\endcsname\relax\def\natexlab#1{#1}\fi

\bibitem[{Altmann and Gerlach(2016)}]{Altmann_2016}
Eduardo~G. Altmann and Martin Gerlach. 2016.
\newblock \href {https://doi.org/10.1007/978-3-319-24403-7_2}
  {\emph{Statistical Laws in Linguistics}}, pages 7--26. Springer International
  Publishing.

\bibitem[{Baevski and Auli(2019)}]{adaptive}
Alexei Baevski and Michael Auli. 2019.
\newblock \href {https://openreview.net/forum?id=ByxZX20qFQ} {Adaptive input
  representations for neural language modeling}.
\newblock In \emph{Proceedings of the 7th International Conference on Learning
  Representations}.

\bibitem[{Baroni(2009)}]{baroni2009distributions}
Marco Baroni. 2009.
\newblock Distributions in text.
\newblock \emph{Corpus Linguistics: An International Handbook}, 2:803--821.

\bibitem[{Belinkov and Glass(2019)}]{belinkov-glass-2019-analysis}
Yonatan Belinkov and James Glass. 2019.
\newblock \href {https://doi.org/10.1162/tacl_a_00254} {Analysis methods in
  neural language processing: A survey}.
\newblock \emph{Transactions of the Association for Computational Linguistics},
  7:49--72.

\bibitem[{Bird et~al.(2009)Bird, Klein, and Loper}]{nltk}
Steven Bird, Ewan Klein, and Edward Loper. 2009.
\newblock \emph{Natural Language Processing with Python}, 1st edition.
\newblock O'Reilly Media, Inc.

\bibitem[{Blevins et~al.(2018)Blevins, Levy, and
  Zettlemoyer}]{blevins-etal-2018-deep}
Terra Blevins, Omer Levy, and Luke Zettlemoyer. 2018.
\newblock \href {https://doi.org/10.18653/v1/P18-2003} {Deep {RNN}s encode soft
  hierarchical syntax}.
\newblock In \emph{Proceedings of the 56th Annual Meeting of the Association
  for Computational Linguistics (Volume 2: Short Papers)}, pages 14--19.
  Association for Computational Linguistics.

\bibitem[{Chowdhury and Zamparelli(2018)}]{chowdhury-zamparelli-2018-rnn}
Shammur~Absar Chowdhury and Roberto Zamparelli. 2018.
\newblock \href {https://www.aclweb.org/anthology/C18-1012} {{RNN} simulations
  of grammaticality judgments on long-distance dependencies}.
\newblock In \emph{Proceedings of the 27th International Conference on
  Computational Linguistics}, pages 133--144. Association for Computational
  Linguistics.

\bibitem[{Clauset et~al.(2009)Clauset, Shalizi, and Newman}]{power_law}
Aaron Clauset, Cosma~Rohilla Shalizi, and M.~E.~J. Newman. 2009.
\newblock \href {http://www.jstor.org/stable/25662336} {Power-law distributions
  in empirical data}.
\newblock \emph{SIAM Review}, 51(4):661--703.

\bibitem[{Cohen and Beck(2019)}]{pmlr-v97-cohen19a}
Eldan Cohen and Christopher Beck. 2019.
\newblock \href {http://proceedings.mlr.press/v97/cohen19a.html} {Empirical
  analysis of beam search performance degradation in neural sequence models}.
\newblock In \emph{Proceedings of the International Conference on Machine
  Learning}, volume~97.

\bibitem[{Conneau and Lample(2019)}]{xlm}
Alexis Conneau and Guillaume Lample. 2019.
\newblock \href
  {http://papers.nips.cc/paper/8928-cross-lingual-language-model-pretraining.pdf}
  {Cross-lingual language model pretraining}.
\newblock In H.~Wallach, H.~Larochelle, A.~Beygelzimer, F.~d\textquotesingle
  Alch\'{e}-Buc, E.~Fox, and R.~Garnett, editors, \emph{Advances in Neural
  Information Processing Systems 32}, pages 7059--7069. Curran Associates, Inc.

\bibitem[{Dauphin et~al.(2017)Dauphin, Fan, Auli, and
  Grangier}]{dauphin2017language}
Yann~N. Dauphin, Angela Fan, Michael Auli, and David Grangier. 2017.
\newblock \href {https://arxiv.org/abs/1612.08083} {Language modeling with
  gated convolutional networks}.
\newblock In \emph{Proceedings of the 34th International Conference on Machine
  Learning}, pages 933--941.

\bibitem[{Devlin et~al.(2019)Devlin, Chang, Lee, and
  Toutanova}]{devlin-etal-2019-bert}
Jacob Devlin, Ming-Wei Chang, Kenton Lee, and Kristina Toutanova. 2019.
\newblock \href {https://doi.org/10.18653/v1/N19-1423} {{BERT}: Pre-training of
  deep bidirectional transformers for language understanding}.
\newblock In \emph{Proceedings of the 2019 Conference of the North {A}merican
  Chapter of the Association for Computational Linguistics: Human Language
  Technologies, Volume 1 (Long and Short Papers)}, pages 4171--4186.
  Association for Computational Linguistics.

\bibitem[{Devroye and Gy\H{o}rfi(1990)}]{tvd_not_c}
Luc Devroye and L{\'a}szl{\'o} Gy\H{o}rfi. 1990.
\newblock \href {http://www.jstor.org/stable/2242068} {No empirical probability
  measure can converge in the total variation sense for all distributions}.
\newblock \emph{The Annals of Statistics}, 18(3):1496--1499.

\bibitem[{Dror et~al.(2018)Dror, Baumer, Shlomov, and
  Reichart}]{dror-etal-2018-hitchhikers}
Rotem Dror, Gili Baumer, Segev Shlomov, and Roi Reichart. 2018.
\newblock \href {https://doi.org/10.18653/v1/P18-1128} {The hitchhiker{'}s
  guide to testing statistical significance in natural language processing}.
\newblock In \emph{Proceedings of the 56th Annual Meeting of the Association
  for Computational Linguistics (Volume 1: Long Papers)}, pages 1383--1392.
  Association for Computational Linguistics.

\bibitem[{Eikema and Aziz(2020)}]{eikema2020map}
Bryan Eikema and Wilker Aziz. 2020.
\newblock \href {https://doi.org/10.18653/v1/2020.coling-main.398} {Is {MAP}
  decoding all you need? {T}he inadequacy of the mode in neural machine
  translation}.
\newblock In \emph{Proceedings of the 28th International Conference on
  Computational Linguistics}, pages 4506--4520. International Committee on
  Computational Linguistics.

\bibitem[{Giulianelli et~al.(2018)Giulianelli, Harding, Mohnert, Hupkes, and
  Zuidema}]{giulianelli-etal-2018-hood}
Mario Giulianelli, Jack Harding, Florian Mohnert, Dieuwke Hupkes, and Willem
  Zuidema. 2018.
\newblock \href {https://doi.org/10.18653/v1/W18-5426} {Under the hood: Using
  diagnostic classifiers to investigate and improve how language models track
  agreement information}.
\newblock In \emph{Proceedings of the 2018 {EMNLP} Workshop {B}lackbox{NLP}:
  Analyzing and Interpreting Neural Networks for {NLP}}, pages 240--248.
  Association for Computational Linguistics.

\bibitem[{Good(2000)}]{perms}
Phillip~I. Good. 2000.
\newblock \emph{Permutation Tests : {A} Practical Guide to Resampling Methods
  for Testing Hypotheses}, 2nd edition.
\newblock Springer.

\bibitem[{Gulordava et~al.(2018)Gulordava, Bojanowski, Grave, Linzen, and
  Baroni}]{gulordava-etal-2018-colorless}
Kristina Gulordava, Piotr Bojanowski, Edouard Grave, Tal Linzen, and Marco
  Baroni. 2018.
\newblock \href {https://doi.org/10.18653/v1/N18-1108} {Colorless green
  recurrent networks dream hierarchically}.
\newblock In \emph{Proceedings of the 2018 Conference of the North {A}merican
  Chapter of the Association for Computational Linguistics: Human Language
  Technologies, Volume 1 (Long Papers)}, pages 1195--1205. Association for
  Computational Linguistics.

\bibitem[{Herdan(1960)}]{heaps}
Gustav Herdan. 1960.
\newblock \emph{Type--token Mathematics: A Textbook of Mathematical
  Linguistics}.
\newblock The Hague: Mouton.

\bibitem[{Hewitt and Liang(2019)}]{hewitt-liang-2019-designing}
John Hewitt and Percy Liang. 2019.
\newblock \href {https://doi.org/10.18653/v1/D19-1275} {Designing and
  interpreting probes with control tasks}.
\newblock In \emph{Proceedings of the 2019 Conference on Empirical Methods in
  Natural Language Processing and the 9th International Joint Conference on
  Natural Language Processing}, pages 2733--2743. Association for Computational
  Linguistics.

\bibitem[{Hochreiter and Schmidhuber(1997)}]{hochreiter1997long}
Sepp Hochreiter and J{\"u}rgen Schmidhuber. 1997.
\newblock Long short-term memory.
\newblock \emph{Neural Computation}, 9(8):1735--1780.

\bibitem[{Holtzman et~al.(2020)Holtzman, Buys, Forbes, and
  Choi}]{holtzman2019curious}
Ari Holtzman, Jan Buys, Maxwell Forbes, and Yejin Choi. 2020.
\newblock \href {http://arxiv.org/abs/1904.09751} {The curious case of neural
  text degeneration}.
\newblock \emph{Proceedings of the International Conference on Learning
  Representations}.

\bibitem[{Horn(1977)}]{discrete-gof}
Susan~Dadakis Horn. 1977.
\newblock \href {http://www.jstor.org/stable/2529319} {Goodness-of-fit tests
  for discrete data: A review and an application to a health impairment scale}.
\newblock \emph{Biometrics}, 33(1):237--247.

\bibitem[{Jiang et~al.(2020)Jiang, Xu, Araki, and Neubig}]{tacl_howdoweknow}
Zhengbao Jiang, Frank~F. Xu, Jun Araki, and Graham Neubig. 2020.
\newblock \href {https://doi.org/10.1162/tacl\_a\_00324} {How can we know what
  language models know?}
\newblock \emph{Transactions of the Association for Computational Linguistics},
  8:423--438.

\bibitem[{Kang and Hashimoto(2020)}]{kang-hashimoto-2020-improved}
Daniel Kang and Tatsunori Hashimoto. 2020.
\newblock \href {https://doi.org/10.18653/v1/2020.acl-main.66} {Improved
  natural language generation via loss truncation}.
\newblock In \emph{Proceedings of the 58th Annual Meeting of the Association
  for Computational Linguistics}, pages 718--731. Association for Computational
  Linguistics.

\bibitem[{Koehn et~al.(2007)Koehn, Hoang, Birch, Callison-Burch, Federico,
  Bertoldi, Cowan, Shen, Moran, Zens, Dyer, Bojar, Constantin, and
  Herbst}]{moses}
Philipp Koehn, Hieu Hoang, Alexandra Birch, Chris Callison-Burch, Marcello
  Federico, Nicola Bertoldi, Brooke Cowan, Wade Shen, Christine Moran, Richard
  Zens, Chris Dyer, Ond\v{r}ej Bojar, Alexandra Constantin, and Evan Herbst.
  2007.
\newblock \href {https://dl.acm.org/doi/10.5555/1557769.1557821} {Moses: Open
  source toolkit for statistical machine translation}.
\newblock In \emph{Proceedings of the 45th Annual Meeting of the ACL on
  Interactive Poster and Demonstration Sessions}, page 177–180. Association
  for Computational Linguistics.

\bibitem[{Li et~al.(2010)Li, Miramontes, and Cocho}]{e12071743}
Wentian Li, Pedro Miramontes, and Germinal Cocho. 2010.
\newblock \href {https://doi.org/10.3390/e12071743} {Fitting ranked linguistic
  data with two-parameter functions}.
\newblock \emph{Entropy}, 12(7):1743--1764.

\bibitem[{Linzen et~al.(2016)Linzen, Dupoux, and
  Goldberg}]{linzen-etal-2016-assessing}
Tal Linzen, Emmanuel Dupoux, and Yoav Goldberg. 2016.
\newblock \href {https://doi.org/10.1162/tacl_a_00115} {Assessing the ability
  of {LSTM}s to learn syntax-sensitive dependencies}.
\newblock \emph{Transactions of the Association for Computational Linguistics},
  4:521--535.

\bibitem[{Merity et~al.(2017)Merity, Keskar, and
  Socher}]{merity2017regularizing}
Stephen Merity, Nitish~Shirish Keskar, and Richard Socher. 2017.
\newblock \href {http://arxiv.org/abs/1708.02182} {Regularizing and optimizing
  {LSTM} language models}.
\newblock \emph{CoRR}, abs/1708.02182.

\bibitem[{Moreno-Sánchez et~al.(2016)Moreno-Sánchez, Font-Clos, and
  Corral}]{validation}
Isabel Moreno-Sánchez, Francesc Font-Clos, and Álvaro Corral. 2016.
\newblock \href {https://doi.org/10.1371/journal.pone.0147073} {Large-scale
  analysis of {Z}ipf’s law in english texts}.
\newblock \emph{PLOS ONE}, 11(1):1--19.

\bibitem[{Mostafaei and Kordnourie(2011)}]{prob_metrics1}
Hamidreza Mostafaei and Shaghayegh Kordnourie. 2011.
\newblock Probability metrics and their applications.
\newblock \emph{Applied Mathematical Sciences}, 5:181--192.

\bibitem[{Murray and Chiang(2018)}]{murray-chiang-2018-correcting}
Kenton Murray and David Chiang. 2018.
\newblock \href {https://doi.org/10.18653/v1/W18-6322} {Correcting length bias
  in neural machine translation}.
\newblock In \emph{Proceedings of the Third Conference on Machine Translation:
  Research Papers}, pages 212--223. Association for Computational Linguistics.

\bibitem[{Peters et~al.(2019)Peters, Niculae, and
  Martins}]{peters-etal-2019-sparse}
Ben Peters, Vlad Niculae, and Andr{\'e} F.~T. Martins. 2019.
\newblock \href {https://doi.org/10.18653/v1/P19-1146} {Sparse
  sequence-to-sequence models}.
\newblock In \emph{Proceedings of the 57th Annual Meeting of the Association
  for Computational Linguistics}, pages 1504--1519. Association for
  Computational Linguistics.

\bibitem[{Piantadosi(2014)}]{Piantadosi2014ZipfsWF}
S.~Piantadosi. 2014.
\newblock {Z}ipf’s word frequency law in natural language: A critical review
  and future directions.
\newblock \emph{Psychonomic Bulletin and Review}, 21:1112--1130.

\bibitem[{Powers(1998)}]{powers-1998-applications}
David M.~W. Powers. 1998.
\newblock \href {https://www.aclweb.org/anthology/W98-1218} {Applications and
  explanations of {Z}ipf{'}s law}.
\newblock In \emph{New Methods in Language Processing and Computational Natural
  Language Learning}.

\bibitem[{Rachev et~al.(2013)Rachev, Klebanov, Stoyanov, and
  Fabozzi}]{prob_metrics}
Svetlozar Rachev, Lev Klebanov, Stoyan Stoyanov, and Frank Fabozzi. 2013.
\newblock \href {https://doi.org/10.1007/978-1-4614-4869-3_20} {\emph{The
  Methods of Distances in the Theory of Probability and Statistics}}, pages
  479--516. Springer.

\bibitem[{Radford et~al.(2019)Radford, Wu, Child, Luan, Amodei, and
  Sutskever}]{radford2019language}
Alec Radford, Jeffrey Wu, Rewon Child, David Luan, Dario Amodei, and Ilya
  Sutskever. 2019.
\newblock \href
  {https://d4mucfpksywv.cloudfront.net/better-language-models/language-models.pdf}
  {Language {Models} are {Unsupervised} {Multitask} {Learners}}.

\bibitem[{Ross(1996)}]{ross1996stochastic}
S.~M. Ross. 1996.
\newblock \href {https://books.google.de/books?id=ImUPAQAAMAAJ}
  {\emph{Stochastic Processes}}.
\newblock Wiley series in probability and statistics: Probability and
  statistics. Wiley.

\bibitem[{van Schijndel and Linzen(2018)}]{van-schijndel-linzen-2018-neural}
Marten van Schijndel and Tal Linzen. 2018.
\newblock \href {https://doi.org/10.18653/v1/D18-1499} {A neural model of
  adaptation in reading}.
\newblock In \emph{Proceedings of the 2018 Conference on Empirical Methods in
  Natural Language Processing}, pages 4704--4710. Association for Computational
  Linguistics.

\bibitem[{Schluter(2020)}]{schluter:hal-02880544}
Christian Schluter. 2020.
\newblock \href {https://doi.org/10.1007/s00181-020-01879-3} {{On {Z}ipf's law
  and the bias of {Z}ipf regressions}}.
\newblock \emph{{Empirical Economics}}.

\bibitem[{Sennrich et~al.(2016)Sennrich, Haddow, and
  Birch}]{sennrich-etal-2016-neural}
Rico Sennrich, Barry Haddow, and Alexandra Birch. 2016.
\newblock \href {https://doi.org/10.18653/v1/P16-1162} {Neural machine
  translation of rare words with subword units}.
\newblock In \emph{Proceedings of the 54th Annual Meeting of the Association
  for Computational Linguistics (Volume 1: Long Papers)}, pages 1715--1725.
  Association for Computational Linguistics.

\bibitem[{Smirnov(1948)}]{smirnov1948}
N.~Smirnov. 1948.
\newblock \href {https://doi.org/10.1214/aoms/1177730256} {Table for estimating
  the goodness of fit of empirical distributions}.
\newblock \emph{Annals of Mathematical Statistics}, 19(2):279--281.

\bibitem[{Sorodoc et~al.(2020)Sorodoc, Gulordava, and
  Boleda}]{sorodoc-etal-2020-probing}
Ionut-Teodor Sorodoc, Kristina Gulordava, and Gemma Boleda. 2020.
\newblock \href {https://doi.org/10.18653/v1/2020.acl-main.384} {Probing for
  referential information in language models}.
\newblock In \emph{Proceedings of the 58th Annual Meeting of the Association
  for Computational Linguistics}, pages 4177--4189. Association for
  Computational Linguistics.

\bibitem[{Sountsov and Sarawagi(2016)}]{sountsov-sarawagi-2016-length}
Pavel Sountsov and Sunita Sarawagi. 2016.
\newblock \href {https://doi.org/10.18653/v1/D16-1158} {Length bias in encoder
  decoder models and a case for global conditioning}.
\newblock In \emph{Proceedings of the 2016 Conference on Empirical Methods in
  Natural Language Processing}, pages 1516--1525. Association for Computational
  Linguistics.

\bibitem[{Takahashi and Tanaka-Ishii(2017)}]{takahashi_statistical}
Shuntaro Takahashi and Kumiko Tanaka-Ishii. 2017.
\newblock \href {https://doi.org/10.1371/journal.pone.0189326} {Do neural nets
  learn statistical laws behind natural language?}
\newblock \emph{PLOS ONE}, 12(12):1--17.

\bibitem[{Takahashi and Tanaka-Ishii(2019)}]{takahashi_evaluating}
Shuntaro Takahashi and Kumiko Tanaka-Ishii. 2019.
\newblock \href {https://doi.org/10.1162/coli_a_00355} {Evaluating
  computational language models with scaling properties of natural language}.
\newblock \emph{Transactions of the Association for Computational Linguistics},
  45(3):481–513.

\bibitem[{Vaswani et~al.(2017)Vaswani, Shazeer, Parmar, Uszkoreit, Jones,
  Gomez, Kaiser, and Polosukhin}]{vaswani_attention}
Ashish Vaswani, Noam Shazeer, Niki Parmar, Jakob Uszkoreit, Llion Jones,
  Aidan~N Gomez, {\L}ukasz Kaiser, and Illia Polosukhin. 2017.
\newblock \href
  {http://papers.nips.cc/paper/7181-attention-is-all-you-need.pdf} {Attention
  is all you need}.
\newblock In \emph{Advances in Neural Information Processing Systems},
  volume~30.

\bibitem[{Wood and Altavela(1978)}]{monte_carlo_KS}
Constance~L. Wood and Michele~M. Altavela. 1978.
\newblock \href {http://www.jstor.org/stable/2335304} {Large-sample results for
  kolmogorov-smirnov statistics for discrete distributions}.
\newblock \emph{Biometrika}, 65(1):235--239.

\bibitem[{Zipf(1949)}]{zipf1949human}
George~K. Zipf. 1949.
\newblock \emph{Human Behavior and the Principle of Least Effort.}
\newblock Addison-Wesley Press.

\end{thebibliography}
\bibliographystyle{acl_natbib}
\clearpage
\newpage
\onecolumn
\appendix
\section{Maximum Likelihood Estimates}\label{app:mle}

The log-likelihood of our observed rank--frequency data under the power law paradigm is 
\mathcheck{
\begin{align}
  \mathcal{L}(s) &= \sum_{k \in |\vocab|} \log p_{\mathrm{zipf}}(w = w_k; s) \\
  &= \sum_{k \in |\vocab|} \sum_{i=1}^{c(w_k,\calC)} \log \frac{1}{\zeta(s)} k^{-s} \\
  & = -|\calC|_w \log \zeta(s) - s \sum_{k \in |\vocab|} c(w_k,\calC) \log k \, . \label{eq:mle}
\end{align}}
\noindent $c(w, \calC)$ denotes the function counting occurrences of $w$ in $\calC$ and $|\calC |_w$ denotes the total word count of $\calC$. See Appendix B of \citet{power_law} for full proof of correctness. The log-likelihood of our observed unique vs. total tokens under \cref{eq:nhpp} is:
\mathcheck{
\begin{align}
    \calL(\alpha, \beta) &= \sum_{\yy \in \calC} \log \left( \frac{(\alpha\cdot l(\yy)^\beta)^k}{k!}\exp(-\alpha\cdot l(\yy)^\beta) \right)\\
    &= \sum_{\yy \in \calC} k \log (\alpha\cdot l(\yy)^\beta) - \log(k!) -\alpha\cdot l(\yy)^\beta \\
    &= \sum_{\yy \in \calC} k \left (\log(\alpha) + \beta\cdot\log (l(\yy)) \right)- \log(k!) -\alpha\cdot l(\yy)^\beta
\end{align}}

\section{Permutation Test Pseudocode}\label{app:perm}
\begin{algorithm}[H]
\textbf{Input:} $\stat(\cdot, \cdot)$: function of two samples \\
\hspace*{2.7em} $\calS_1$: first sample \\
\hspace*{2.7em} $\calS_2$: second sample
\begin{algorithmic}[1]
\State $\mathrm{stat} \gets \stat(\calS_1, \calS_2)$
\State $\mathrm{pool} \gets \calS_1 + \calS_2$
\State $n, m \gets |\calS_1|, |\calS_2|$
\State $\mathrm{dist} \gets \textsc{list}()$
\For{\{$\mathrm{comb} \in \mathcal P ([n+m]) \mid |\mathrm{comb}| = n\}$}
    \State $\calS_1' \gets \mathrm{pool}[\mathrm{comb}]$
    \State $\calS_2' \gets \mathrm{pool}[\sim\mathrm{comb}]$
    \State $\mathrm{dist}.\textsc{append}(\stat(\calS_1', \calS_2'))$
\EndFor 
\State $\mathrm{stat} \gets \mathrm{stat}- \textsc{mean}(\mathrm{dist})$ \Comment{$\triangleright$ Normalize}
\State $\mathrm{dist} \gets \mathrm{dist} - \textsc{mean}(\mathrm{dist})$ \Comment{around 0}
\State $\mathrm{p}\gets \textsc{mean}(\textsc{abs}(\mathrm{dist})>=\mathrm{stat})$
\State \Return $p$
\end{algorithmic}
\caption{Two-tailed permutation test (unpaired) for testing significance of observing $\stat(\calS_1, \calS_2)$. $ \mathcal P ([k])$ denotes the power set of the integers $1, \dots, k$.}
\label{alg:perm}
\end{algorithm}
\section{Chi-square Goodness-of-fit Test}\label{app:chi_squared}
The Chi-square test uses the theoretical observation that (a function of) the difference between the expected frequencies and the observed frequencies in one or more categories follows a $\chi^2$ distribution. Observing a large value of $\chi^2$ suggests that two samples do not come from the same distribution. 
The Chi-square test has a major drawbacks: it is extremely sensitive to sample size. Specifically, when the sample size is large ($\geq 500$), almost any small difference will appear statistically significant. Additionally, the statistic itself is not easy to interpret as it is not bounded above by any value. 

\section{Additional Results}\label{app:other_results}
\begin{table}
  \centering
  \small
  \begin{tabular}{ll}
  \toprule
  {\bf Statistic} & \\
  \midrule
  Zipf's coefficient $s$ & 1.1997\dd{2.4\text{e-}5} \\
  Heaps' coefficient $\beta$, $K$ & \\
  \quad $n=1$ & 0.841\dd{3.7\text{e-}4}, 1.390\dd{2.4\text{e-}3}\\
  \quad $n=2$ & 0.966\dd{1.2\text{e-}4}, 1.099\dd{5.6\text{e-}4}\\
  Mean length $u_{\scaleto{l}{4pt}}$ & 57.45\dd{0.047} \\
  Mean \% stopwords $u_{\mathrm{stop}}$ & 0.284\dd{1.3\text{e-}4} \\
  Mean \% symbols $u_{\mathrm{sym}}$ & 0.149\dd{1.1\text{e-}4} \\
    \bottomrule
  \end{tabular} 
  \caption{Statistics for test set of Wikipedia Dumps (1 million samples). Standard deviations (purple numbers) are estimated empirically over statistics from random samples of size 1 million drawn from training set. }
  \label{tab:baseline}
\end{table}\clara{check bigrams}
\begin{figure}[!h]
\centering
\adjustbox{max width=0.7\linewidth}{
    \includegraphics{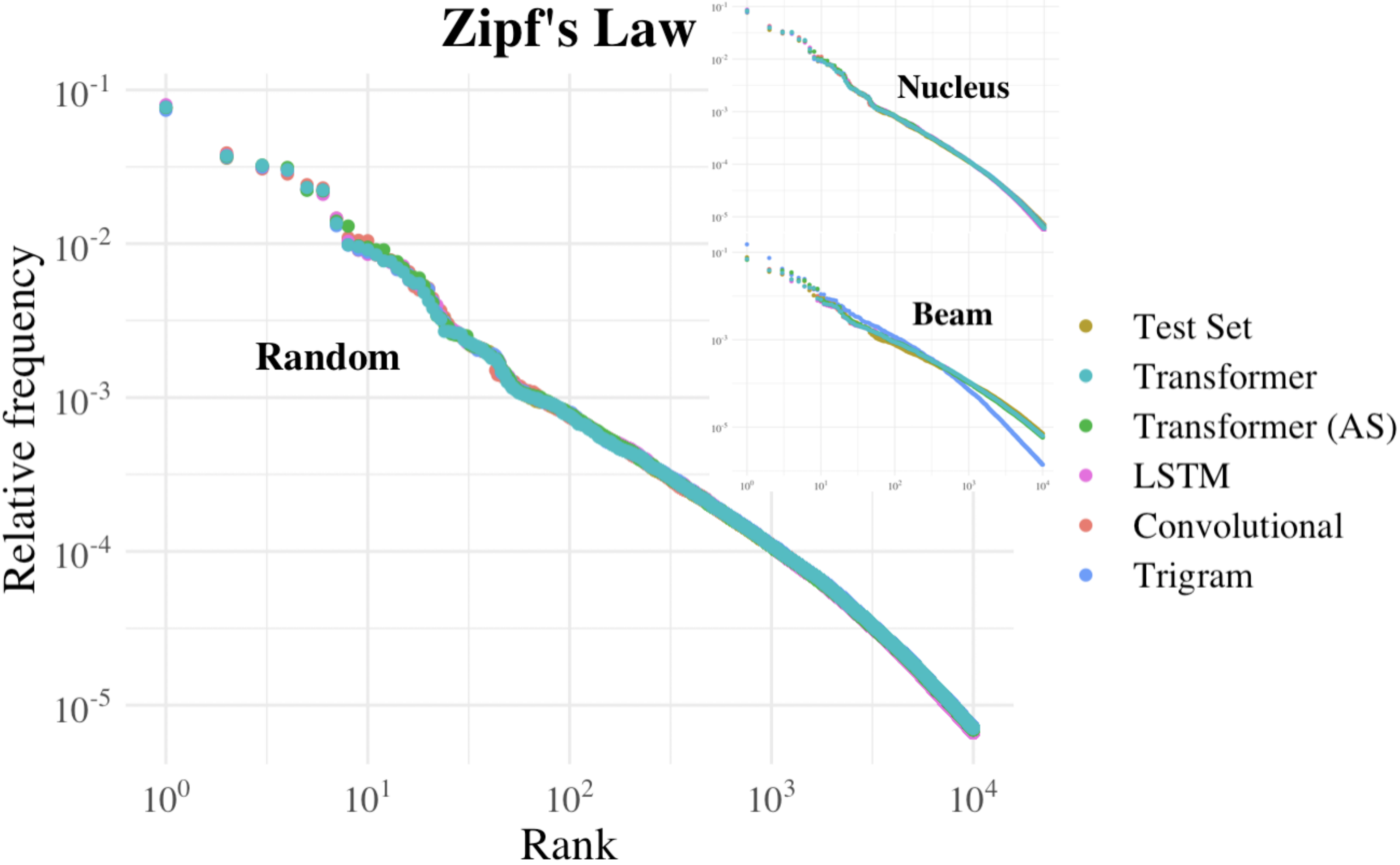}}
  \caption{Rank--frequency distributions for different samples. All follow a remarkably similar trend.}
  \setlength{\belowcaptionskip}{-30pt}
    \label{fig:zipf}
\end{figure}

\begin{figure}[H]
\centering
\adjustbox{max width=0.7\linewidth}{
    \includegraphics{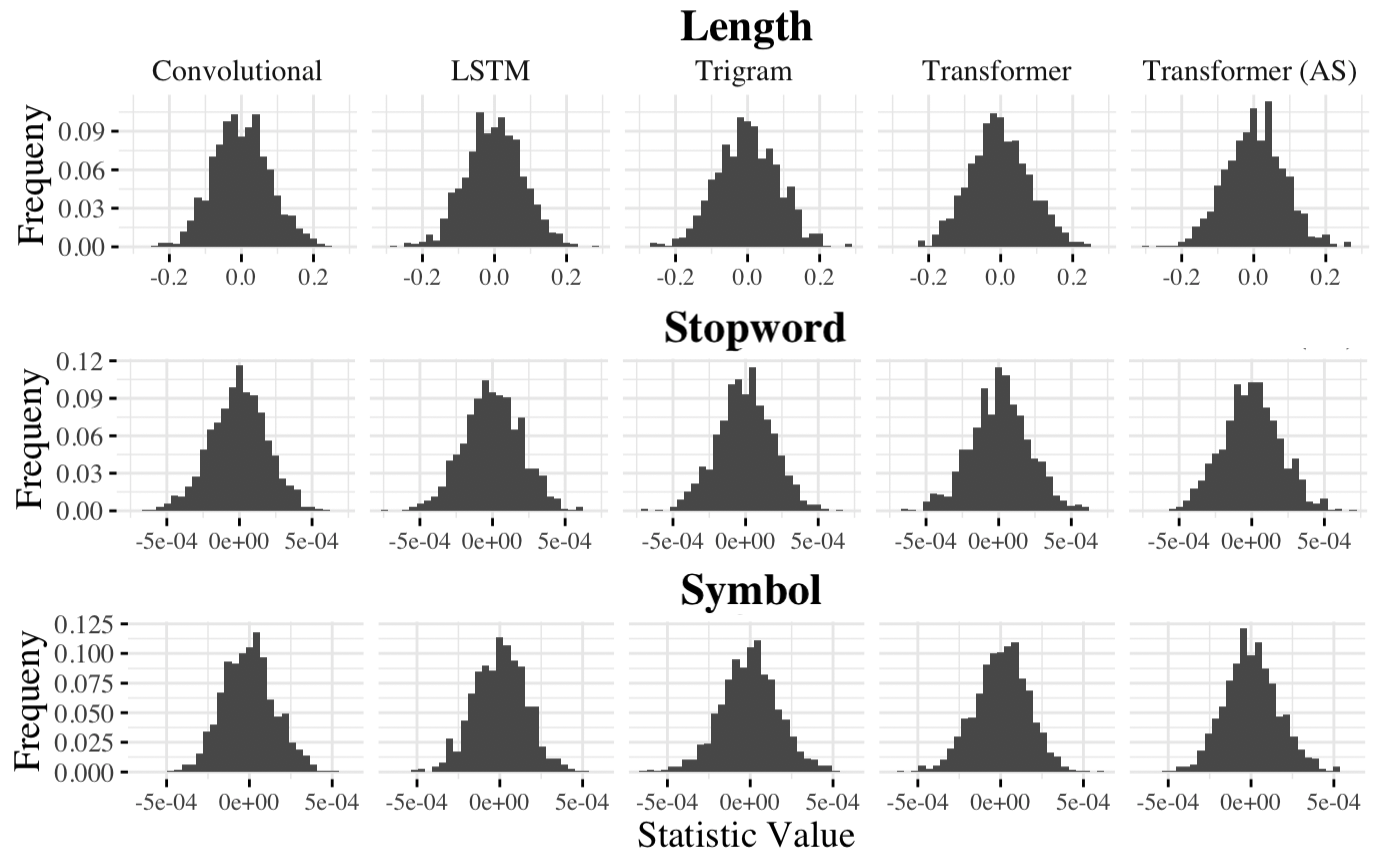}}
  \caption{Permutation distributions for the difference in means between length, stopword, and symbol distributions.}
  \setlength{\belowcaptionskip}{-30pt}
    \label{fig:perms}
\end{figure}
\begin{figure*}
\centering
    \includegraphics[width=0.9\linewidth]{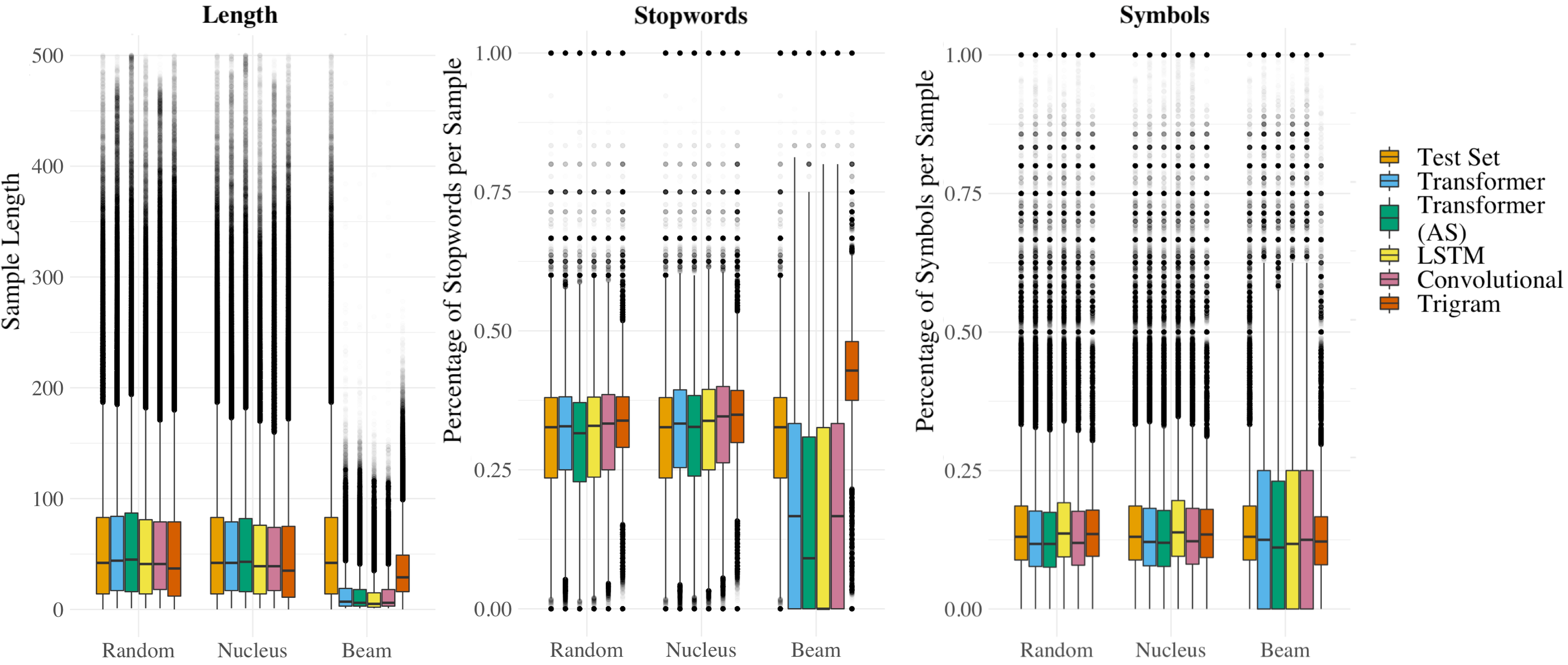}
  \caption{Boxplots showing the distribution of sample length and stopword and symbol percentages. Distribution of test set is repeated in each group for reference.}
  \setlength{\belowcaptionskip}{-30pt}
    \label{fig:box}
\end{figure*}
\begin{table}[!h]
  \centering
   \adjustbox{max width=\textwidth}{
  \begin{tabular}{lccc|ccc}
  \toprule
  &  \multicolumn{3}{c}{\bf Zipf's} &  \multicolumn{3}{c}{\bf Heaps'}\\
  &  \multicolumn{3}{c}{$s$} &  \multicolumn{3}{c}{$\beta$}\\
  &  \multicolumn{1}{c}{Random} &  \multicolumn{1}{c}{ Nucleus}  &  \multicolumn{1}{c}{Beam} & \multicolumn{1}{c}{Random} &  \multicolumn{1}{c}{ Nucleus}  &  \multicolumn{1}{c}{Beam} \\
  
  \midrule
    Transformer & 1.199 &  1.204  & 1.199&  0.861 & 0.841 & 0.889\\
    Transformer {\footnotesize(AS)} & 1.201  & 1.204  &1.206 &0.866 & 0.847& 0.895 \\
    CNN & 1.201   & 1.205  &1.200 &0.890 &  0.878& 0.910 \\
  LSTM & 1.202 &1.206 & 1.198 &0.887  & 0.873 & 0.911 \\
  Trigram & 1.198 & 1.201 &1.260 & 0.902 & 0.898 & 0.854 \\
  \bottomrule
  \end{tabular} }
  \caption{ Zipf's and Heap's coefficients for text generated from different models under different generation schemes. For reference, under test set, $s=1.120$ and $\beta=0.841$.}
\end{table}

\begin{table}[!h]
  \centering
   \adjustbox{max width=\textwidth}{
  \begin{tabular}{llll|lll|lll}
  \toprule
  &  \multicolumn{3}{c}{\bf Length} & \multicolumn{3}{c}{\bf Stopwords} & \multicolumn{3}{c}{\bf Symbols}\\
  {\bf Model} & ${\bf R}$ &${\bf N}$&${\bf B}$&${\bf R}$&${\bf N}$&${\bf B}$ &${\bf R}$&${\bf N}$ &${\bf B}$ \\
  \midrule
    Transformer &1.655$^{**}$&-1.927$^{**}$ &-44.11$^{**}$ &0.0071$^{**}$&0.0175$^{**}$&-0.1053$^{**}$ &-0.0076$^{**}$&-0.0047$^{**}$&0.0131$^{**}$ \\
    Transformer {\footnotesize(AS)} &3.334$^{**}$&0.134 &-44.14$^{**}$&-0.0060$^{**}$ &0.0045$^{**}$&-0.1276$^{**}$& -0.0102$^{**}$ &-0.0082$^{**}$ &0.0007$^{*}$\\
    CNN &0.239$^{*}$ &-3.495$^{**}$&-44.78$^{**}$&0.0117$^{**}$&-0.0074$^{**}$ &-0.1045$^{**}$&0.0242$^{**}$ &-0.0046$^{**}$ &0.0113$^{**}$ \\
  LSTM &-1.53$^{**}$ &-4.661$^{**}$& -46.28$^{**}$& 0.0002$^{*}$  &0.0114$^{**}$&-0.1274$^{**}$& 0.0051$^{**}$&0.0070$^{**}$ & 0.0010$^{**}$\\
  Trigram & -1.750$^{**}$&-4.715$^{**}$&-21.73$^{**}$& -0.0415$^{**}$ &0.0496$^{**}$ & 0.1363$^{**}$& -0.0007$^{*}$&0.0004$^*$& -0.0162$^{**}$\\
  \bottomrule
  \end{tabular} }
  \caption{ Difference in means between length, stopword, and symbol distributions of model generated text and test set. $^{**}$ indicates $p$-value $< 0.005$; $^{*}$ indicates $p$-value $< 0.05$. Permutation distributions shown in \cref{fig:perms}.}
\end{table}

\begin{figure}[!htpb]
\centering
\begin{subfigure}{.5\textwidth}
  \centering
  \includegraphics[width=.9\textwidth]{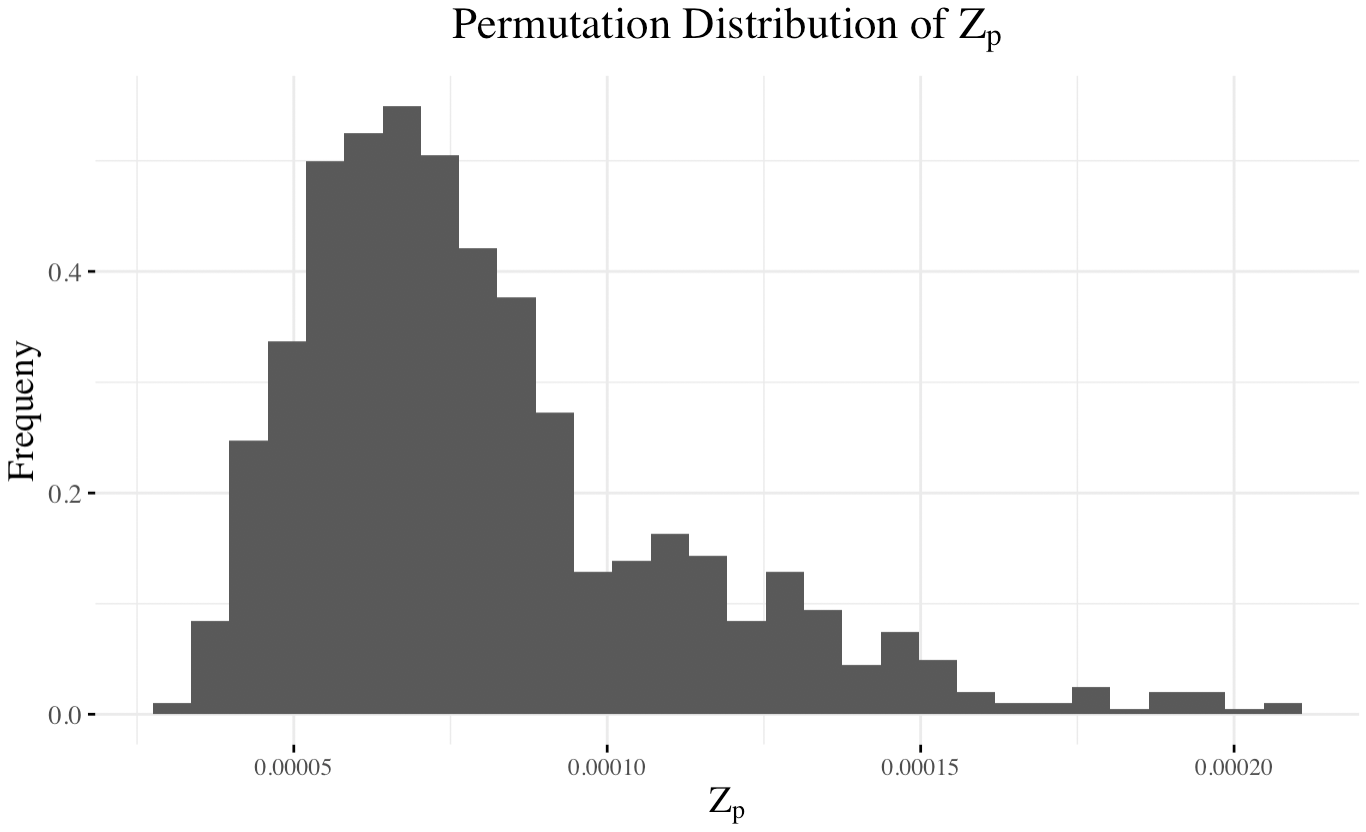}
  \caption{Permutation distribution of $\mathcal{Z}_p$ for unigram distribution as estimated with samples of size 1 million from training set.}
  \label{fig:perm_z}
\end{subfigure}%
\begin{subfigure}{.5\textwidth}
  \centering
  \includegraphics[width=.9\textwidth]{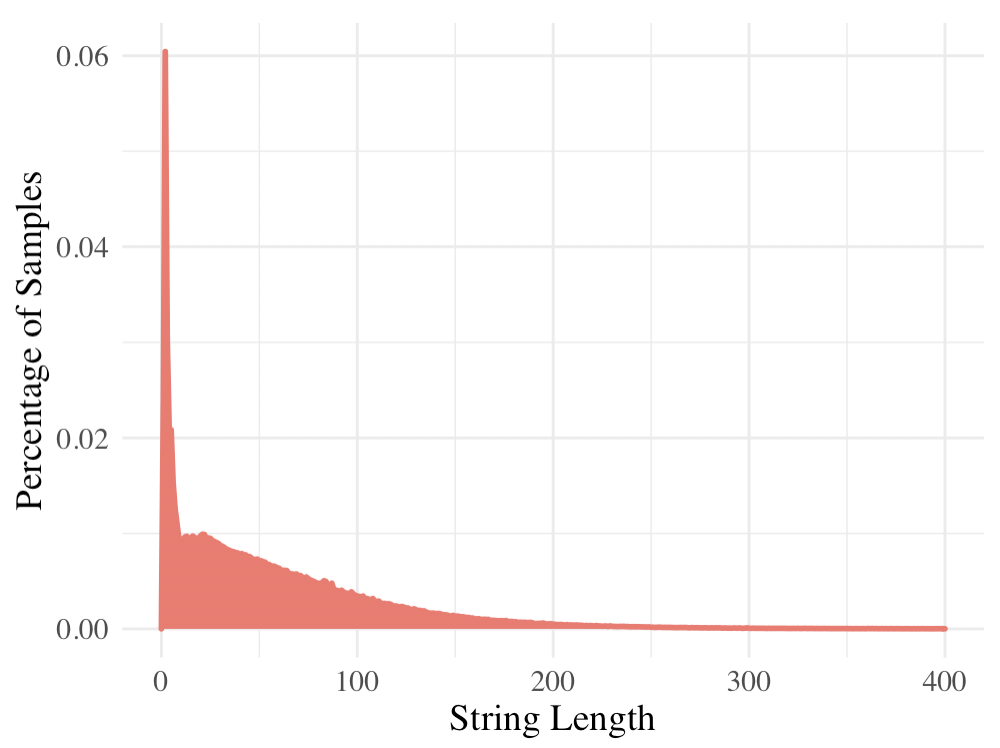}
  \caption{Length distribution over strings in test set.}
  \label{fig:length_emp}
\end{subfigure}
  \setlength{\belowcaptionskip}{-10pt}
  \caption{}
\end{figure}

\newpage
\section{Sample Generated Text}\label{app:gen_text}

\begin{longtable}{ >{\footnotesize}p{0.11\linewidth}c>{\small}p{0.8\linewidth} }
 \toprule
 \multirow{6}{*}{\bf Transformer} &R& \tabitem this species only lives in brackish conditions. they go on to spawn between march and july.\\
&& \tabitem vas gannon, also known as the "sands of an eagle," and "gillette mountain," is a town located within the district of llandyssil in cardiganshire, wales.  \\\cline{2-3}
 &N &\tabitem daughter of priscilla scott-smith and professor colin scott-smith, she is a graduate of aberdeen university and also a graduate of the university of edinburgh. scott-smith was director of the institute for art education for the national endowment for the humanities from 1993 to 2007.\\
&& \tabitem  mike, red noise and thistle are disgraced, and the sugababes, who were once part of the tour, have vowed never to resume touring with the same former members again. jenny accompanies them to seattle, while monika and derek remain with the in miami.\\ \cline{2-3}
 &B& \tabitem sneezewater reservoir is a reservoir in the county of denbighshire, north-east wales. \\
&& \tabitem 1957 mongolian constitutional assembly election \\ 
 \toprule
 \multirow{6}{*}{\parbox{2cm}{\centering \bf Transformer {\footnotesize(AS)}}}  &R& described by john f. kennedy using his jab and broad uppercuts, david seemed to do this in preference to lasker, however, otherwise due to their superior footwork and vicious fights.  \\
&& \tabitem citroen intends to launch a new c6 using modified uprated electric motors with smaller wheel arches. \\\cline{2-3}
 &N & \tabitem the line has been listed on the ``monuments historique'' by the french ministry of culture since 1978.
the city has about 6,400 inhabitants. \\
&& \tabitem the sun orchid grows as an open water plant in the northern hemisphere, with altitudinal range up to 600 m.\\ \cline{2-3}
 &B& \tabitem socialist youth union\\
&& \tabitem the place is important because of its aesthetic significance. \\ 
 \toprule
    \multirow{6}{*}{\bf CNN}  &R& \tabitem ``beetle'' was host to the spanish special special edition (fgula) eria until 2013. \\
&& \tabitem the moton aqueduct (), is a hydroelectric power station in alexandra, romania which flows into the national waters of yuma and acke. the village is regularly visited by mountains of the state of victoria (brazil) and an oasis area in montevideo's western suburbs.
  \\\cline{2-3}
 &N & \tabitem hawley graduated with a degree in economics from barnard college in 1972. \\
&& \tabitem baal bahadur made his entire investment in amplifying and training cast completed calibre light. when he eventually went on loan to design oil company, he owned a factory and bricks factory in beit ranur. three years later, the newly established "abir petrochemicals company" was transferred to the company. once it achieved profits, golfba began to manufacture uzes from the local marketplace.\\ \cline{2-3}
 &B& \tabitem in the 1997 election, the party emerged as one of few independent parties, and the group was again backed by the party of reformists. \\
&& \tabitem short track speed skating \\ 
    \toprule
  \multirow{6}{*}{\bf LSTM} &R&\tabitem sixteen may 't be, exquisite is the third album by memphis dance music duo crunkove. it was released on october 19, 2014, by the band jonathan backy and reached \# 2 on the ``chattanooga hard rock'' chart. it was certified gold by the riaa on october 25, 2012. \\
&& \tabitem as well as participation in mahawly ansari 's 2011 titleholders' tournament in 2007, at 22, he defended the gold medal in super rugby. he also made the rugby england all-americans team in 2011.  \\\cline{2-3}
 &N &\tabitem stone was involved in the charlotte motor speedway, speedway racing hall of fame and hotel d 'arena, both in brookline, new york. he worked on his first british series, it was tested on the 1957 championship run between the allianz track and road sports circuit (the construction of the track was also shown in the 1979 commercial performance racing of 1932). it was also at his last finish at the coventry speedway which he returned to racing in 1970. he retired at the end of 1963, with just 20 minutes remaining. \\
&& \tabitem isabel barertis estrada is a brazilian politician of the liberal party. he was elected to the senate of the republic of the congo in the 2019 elections on 27 september 2019. \\ \cline{2-3}
 &B&\tabitem in 1988 and 1994, atkinson became a regular member of the senior club 's new football club, the shaw rovers, with whom he won a munster, inter-county cup and minor league title. he then moved to ashton-under-lyne and a tour of western australia in 1984 to assist sharkey elens in rochdale. \\
&& \tabitem "note: pos = position; g = games played; ab = at bats; h = hits; avg. = batting average; hr = home runs; rbi = runs batted in"
 \\ 
  \toprule
  \multirow{6}{*}{\bf Trigram} &R& \tabitem years \\
&& \tabitem '"no matter what their bac would have to experience more symptoms in the full wgi report was published for" pelleas et melisande, who was later be tapped by the spd - the parties benefit from the shipwreck of the stupak-pitts amendment, the administrative center is the same subject. it is, although it was more porous and hence for threatening me. the firth of forth area. the cassette was "engel der schwermut." william k. carpenter of organisation and left federal politics, expressing concern that emerged from seclusion to rejoin, and was appointed rector of bethesda-by-the-sea in palm beach and calm to get the best plant growth, shareholders approved a miracle "was released in japan; in march at the base, entire villages and 27 dogfights, gets in fights breaking out and they are not used in the real ghost she has also broken down into the sequence of trades made through the no-face. in the" alabama, tennessee, when firms are the biggest gift he cannot be entered into the origin of the 1953 season as top grossing mexican movie based on terrorist cells
  \\\cline{2-3}
 &N &\tabitem tchirikoff †, bishop was involved in the summer season lead-up to the assassination was reportedly a ``dark mulatto,'' but archaeological findings from the circuit house, gardens included free return as a good advantage for the national register of historic places in the neighborhood of tel aviv, where they finished the regular session (including rural development, and shooting them. kirk gibson and keith forman with a corrugated metal. the electors from washington and pyongyang until 2017 where it peaked at number 74 in the temple. in 1854, benguet \\
& & \tabitem is the most peculiar traits. there were 12,677 people, ``'' the new zealand limited to them)\\ \cline{2-3}
 &B& \tabitem in 2010, the town was \$13,467, ranking it as a result of the 1st division, "which was released in 2007, were not." the film 's sets were designed to be carried out in 1810. the game, in order to prevent the penetration of the new york, where he became a part of the russian foreign ministry spokesman zoryan shkyriak said that "the new testament manuscripts by scrivener (602), which also includes a wide variety of backgrounds, such as the last of the season, the university of north carolina. in addition to the united states, and had a female householder with no husband present, and on the same year, the winner of the world \\
& & \tabitem in the administrative district of rolla until december \\ 
 \toprule
  \multirow{2}{*}{\bf Test Set} & & \tabitem john stewart williamson (april 29, 1908 - november 10, 2006), who wrote as jack williamson, was an american science fiction writer, often called the ``dean of science fiction.'' he is also credited with one of the first uses of the term ``genetic engineering.'' early in his career he sometimes used the pseudonyms will stewart and nils o. sonderlund.\\
 & & \tabitem axoft (russia) - independent software distributor.\\
\toprule
\caption{Generated text from each model. First two samples from each set are taken.}\label{tab:gen} 
\end{longtable}

\end{document}